\documentclass[accepted]{uai2023} 

\usepackage[dvipsnames]{xcolor}

\usepackage[american]{babel}
\usepackage{amsthm}
\usepackage{natbib} 
\usepackage{mathtools} 
\usepackage{booktabs} 
\usepackage{tikz} 

\usepackage{bbm}
\usepackage{amsfonts, amsmath, amssymb}
\usepackage{algorithm}
\usepackage{algorithmic}
\usepackage{caption}
\usepackage{subcaption}
\hypersetup{
    colorlinks=true,
    allcolors=NavyBlue
}

\newcommand{\x}{\mathbf{x}}

\newcommand{\Ncal}{\mathcal{N}}
\newcommand{\Xcal}{\mathcal{X}}
\newcommand{\norm}[1]{\left\lVert#1\right\rVert}

\newcommand{\R}{\mathbb{R}}

\newcommand{\X}{\mathcal{X}}
\newcommand{\Y}{\mathcal{Y}}

\newcommand{\F}{\mathcal{F}}

\newcommand{\hF}{\hat{F}}
\newcommand{\hp}{\hat{p}}
\newcommand{\indicator}[1]{\mathbbm{1}\{#1\}}
\newcommand{\qedwhite}{\hfill \ensuremath{\Box}}
\DeclareMathOperator*{\argmin}{arg\,min}

\newcommand{\ops}{\text{OPS}}
\newcommand{\mw}{\text{MW}}
\newcommand{\iw}{\text{IW}}
\newcommand{\Ycal}{\mathcal{Y}}
\newcommand{\sigmoid}{\text{sigmoid}}
\newcommand{\logit}{\text{logit}}
\newcommand{\Dcal}{\mathcal{D}}

\newtheorem{definition}{Definition}
\theoremstyle{definition}

\title{Parity Calibration}

\author[1]{\href{mailto:<youngsec@cs.cmu.edu>?Subject=Parity Calibration}{Youngseog Chung}{}}
\author[1]{Aaron Rumack}
\author[1]{Chirag Gupta}
\affil[1]{%
    Machine Learning Department,
    Carnegie Mellon University,
    Pittsburgh, Pennsylvania, USA
}
  
\begin{document}

\hypersetup{urlcolor=black}
\maketitle
\hypersetup{urlcolor=NavyBlue}

\begin{abstract}
In a sequential regression setting, a decision-maker may be primarily concerned with whether the future observation will increase or decrease compared to the current one, rather than the actual value of the future observation. In this context, we introduce the notion of parity calibration, which captures the goal of calibrated forecasting for the increase-decrease (or ``parity") event in a timeseries. Parity probabilities can be extracted from a forecasted distribution for the output, but we show that such a strategy leads to theoretical unpredictability and poor practical performance. We then observe that although the original task was regression, parity calibration can be expressed as binary calibration. Drawing on this connection, we use an online binary calibration method to achieve parity calibration. We demonstrate the effectiveness of our approach on real-world case studies in epidemiology, weather forecasting, and model-based control in nuclear fusion.
\end{abstract}

\section{Introduction}
\label{sec:introduction}
Many tasks in the scope of prediction and decision making are sequential in nature.
A weather forecaster who uses some procedure to make predictions for tomorrow, may find that tomorrow's events falsify these predictions. A good forecaster must then update their model before using it on the following days. In this paper we study the sequential forecasting setting where the goal is to make predictions about a sequence of real-valued outcomes $y_1, y_2, \ldots \in \Y \subseteq \R$ using informative covariates $\x_1, \x_2, \ldots \in \Xcal$. In the presence of inherent stochasticity or insufficient data, forecasters who provide rich predictions in the form of complete distributions over the output allow us to reason about the inherent uncertainties in the data stream \citep{gneiting2007probabilistic}. If a distributional prediction is available, a downstream decision-maker can account for risks that were unknown at the time of forecasting.

Often, a distributional forecast for the real-valued $y_t$ takes the form of 
a predictive cdf (cumulative distribution function) for $y_t$, which in this paper we typically denote as $\hF_t : \mathcal{Y} \to [0,1]$. We sometimes write $\hF_t$ as  $\hF_t(\cdot|\x_t)$ or $\hF_t(\cdot|\x_t, y_{t-1}, \x_{t-1}, \ldots, y_1, \x_1)$; this overloaded notation allows us to be succinct when defining what it means for $\hF_t$ to be calibrated, but explicit when it is necessary to stress that $\hF_t$ depends on all available knowledge. We also refer to $\hF_t$'s as regression forecasts, as it models a continuous distribution over the real-valued output.

In this paper, we are interested in the question: 
can we forecast whether the future outcome $y_{t+1}$ will be greater or less than the current outcome $y_{t}$? To motivate this question, consider a hospital in the midst of a fast moving pandemic such as COVID-19. It may be difficult for the hospital to comprehend absolute numbers of patients requiring hospitalization. However, relative numbers are perhaps easier to interpret: hospitals know the situation today, and would like to know if it is going to worsen or improve tomorrow.

A domain expert (e.g. epidemiologist) may have produced a regression forecast $\hF_t$ for $y_t$. The downstream user (e.g. hospital) can then extract from $\hF_t$ a natural implied probability of the next observation decreasing: 
\begin{equation}
    \text{ for $ t \geq 2$, }\ \hp_t = \hF_t(y_{t-1} \mid \x_t).  \label{eq:naive-strategy}
\end{equation}
The hope of the hospital is that the forecasted probabilities $\hp_t$ are parity calibrated, as defined next. 

\begin{definition}[Parity calibration]
The forecasts $\{\hp_t \in [0,1]\}_{t= 2, \dots, T}$ are said to be parity calibrated if
\begin{equation} \label{eq:parity-calibration-defn}
    \frac{\sum_{t=2}^{T} \indicator{y_t \leq y_{t-1}}\indicator{\hp_t=p}}{\sum_{t=2}^{T} \indicator{\hp_t=p}} \rightarrow p, \forall p \in [0, 1].
\end{equation}
\end{definition}
In words, whenever a parity calibrated forecaster predicts with probability $p$ that $y_t \leq y_{t-1}$, the event $\indicator{y_t \leq y_{t-1}}$ actually occurs with empirical frequency $p$ (in the long run).
To avoid confusion with usage of the term ``parity" in fairness literature, we remark that our context is purely in comparing two consecutive values.

Our first contribution is showing that even if $\hF_t$ is calibrated (based on some accepted notions of calibration), the seemingly reasonable strategy mentioned above \eqref{eq:naive-strategy} can have devastating and unpredictable behavior (Section~\ref{sec:counter-examples}). Yet, it stands to reason that the expert's rich forecast $\hF_t$ should be used in some way. Our second contribution is a methodology for doing this (Sections \ref{methods} and \ref{sec:experiments}). Our main methodology described in Section~\ref{sec:main-parity-calibration-methodology} is based on the key observation that although the parity calibration problem is derived from a regression problem, it naturally reduces to a problem of forecasting binary events.

\ifx false 
We use the term ``parity’’ to denote the task of comparing two consecutive observations, and study the problem of assessing parity from expert forecasts, e.g. \textit{will the next observation be lower than the current value, and what is its (predicted/true) probability?}
W.o.l.g, we can regard the event `` next observation $\leq$ current observation'' as the positive class outcome, and denote the predicted probability attributed to this event as ``parity probability''.
A decision maker would desire the model's parity probabilities to be \textit{calibrated}, and we coin the term ``parity calibration’’ to describe the state of a forecast whose predicted probabilities for parity are aligned with the observations:
out of all instances when a parity calibrated model predicts $p$, the next observation will actually decrease $p$ percent of the time, for any $p \in (0, 1)$.

However, in designing the expert forecast, 
the domain expert may be ignorant of how their forecast will be used downstream, and in learning their model, optimize other notions of ``goodness of fit'', e.g. likelihood, various proper scoring rules, or standard metrics for calibrated regression, including quantile calibration~\citep{gneiting2007probabilistic} or distribution calibration~\citep{song2019distribution}. 
This gives rise to the question, do calibrated regression models ensure parity calibration 
(w.r.t the transformed classification problem)?

We formalize the aforementioned notions in the next section and follow up with this link between calibration metrics in regression and parity calibration.
\fi 

\ifx false \subsection{Problem and notation}
We consider the sequential prediction setting where the goal is to forecast a sequence of real-valued outcomes $y_1, y_2, \ldots \in \Y \subseteq \R$ using informative covariates $\x_1, \x_2, \ldots \in \Xcal$. In its most general form, a forecast for $y_t$ takes the form of 
a predictive cdf (cumulative distribution function) for $y_t$, which we denote as $\hF_t \equiv \hF_t(\cdot|\x_t) \equiv \hF_t(\cdot|\x_t, y_{t-1}, \x_{t-1}, \ldots, y_1, \x_1)$. This overloaded notation for $\hF_t$ allows us to be succinct when defining what it means for $\hF_t$ to be calibrated, but explicit when it is necessary to stress that $\hF_t$ depends on all available knowledge. 

Parity calibration, on the other hand, transforms the problem into classification via the notion of parity between two consecutive observations: 
given that $\hF_t(y_{t-1})$ is the parity probability for time $t$ (i.e. the predicted probability the forecaster effectively assigns to the event $\{y_t \leq y_{t-1}\}$), we assess how aligned it is to 
the observed outcome $\indicator{y_t \leq y_{t-1}}$.
\fi 

\subsection{Regression calibration does not give parity calibration}
\label{sec:counter-examples}
A popular notion of calibration in regression is \textit{probabilistic calibration} \citep{gneiting2007probabilistic}. 
The sequence $\hF_1, \hF_2, \dots$ is said to be probabilistically calibrated if 
\begin{align} \label{eq:probabilistic-calibration}
    \frac{1}{T}\sum_{t=1}^{T}F_t(\hF_t^{-1}(p)) \rightarrow p,\ \forall p \in [0, 1], 
\end{align}
where $F_t$ denotes the ground truth distribution.
Probabilistic calibration is also referred to as \textit{quantile calibration}, since it focuses on the quantile function being valid. In other works, it has also been referred to as average calibration \citep{zhao2020individual, chung2021beyond, sahoo2021reliable}, 
or simply calibration \citep{kuleshov2018accurate, cui2020calibrated, charpentier2022natural, marx2022modular}.
We will henceforth refer to this notion as \textit{quantile calibration}.

Another notion of calibration in regression is \textit{distributional calibration} \citep{song2019distribution}, 
which assesses the convergence of the full distribution of the observations to the predictive distribution. 
A distribution calibrated forecaster satisfies $\forall p \in [0, 1],\ \forall F \in \F$,
\begin{align} \label{eq:distribution-calibration}
    \frac{\sum_{t=1}^{T}\indicator{\hF_t = F} F_t(\hF_t^{-1}(p))}{\sum_{t=1}^{T}\indicator{\hF_t = F}} \rightarrow p,  
\end{align}
where $\F$ is the space of distributions predicted by $\hF_t$.
However, distributional calibration is an idealistic notion that cannot be achieved in practice~\citep{song2019distribution}. 

Recently, \citet{sahoo2021reliable} paired calibration with the notion of threshold decisions and proposed \textit{threshold calibration}.
Forecasts are said to be threshold calibrated if, 
\begin{align*}
    &\frac{\sum_{t=1}^{T}\indicator{\hF_t(y_0) \leq \alpha} F_t(\hF_t^{-1}(p))}{\sum_{t=1}^{T}\indicator{\hF_t(y_0) \leq \alpha}} \rightarrow p, \\
    &\forall y_0 \in \Y,\ \forall \alpha \in [0, 1], \forall p \in [0, 1].
\end{align*}
\citet{sahoo2021reliable} show that distribution calibration implies threshold calibration, but the converse may not hold.

A common aspect of the aforementioned notions of calibration is that they all assess how well-aligned the predictive quantiles are to their empirical counterparts. The key difference among the notions is the conditioning over which this assessment is performed.

Since calibration is regarded as a desirable quality of distributional forecasts, one may wonder whether a calibrated $\hF_t$ is sufficient for parity calibration of the implied probabilities as per Eq. \eqref{eq:naive-strategy}.
We show that this is \textit{not} the case with the following examples. 

\begin{figure}[t]
\vskip 0.2in
\centering 
\includegraphics[width=1.03\columnwidth]
{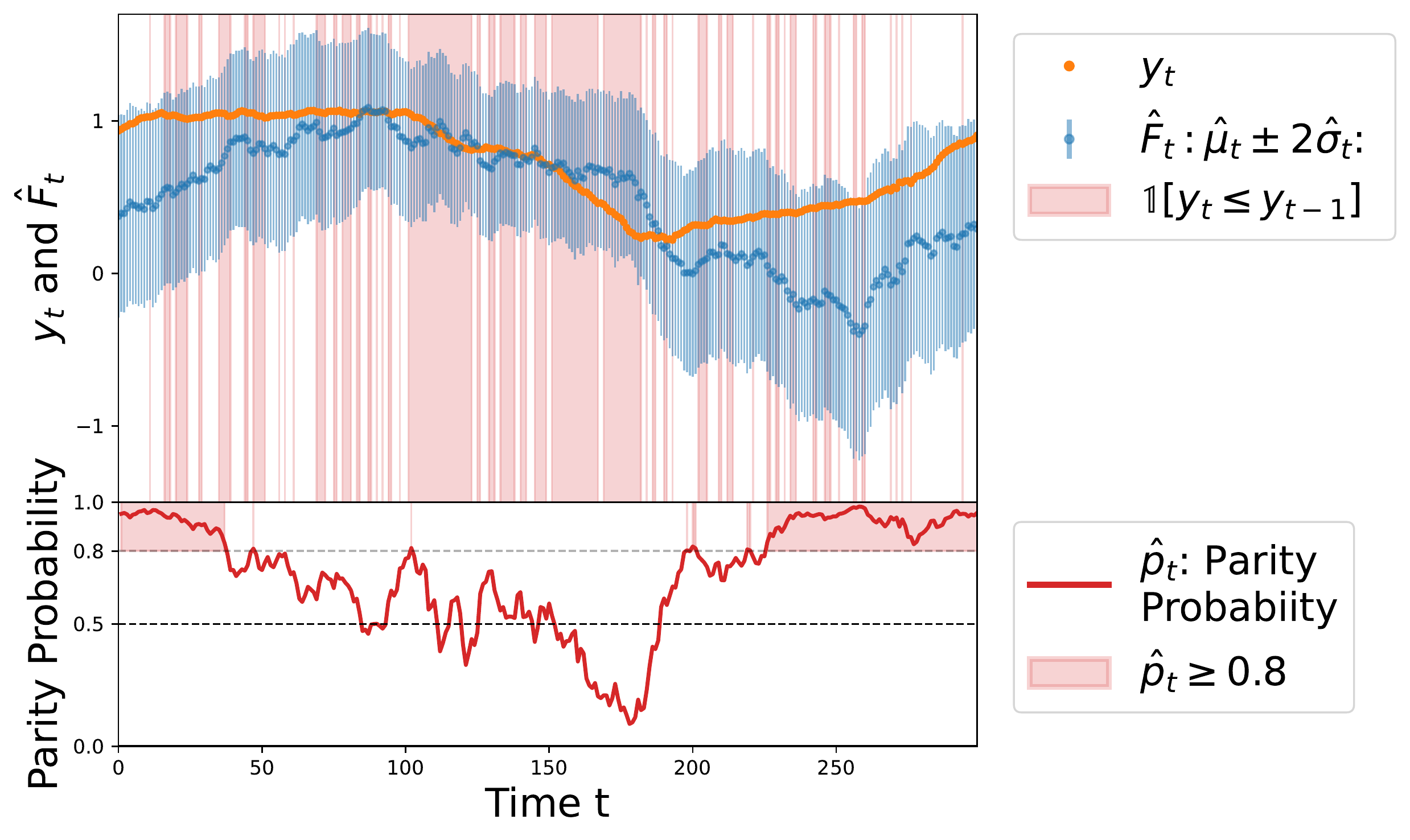}
\caption{Snapshot of the first 300 points from one of our experiment datasets (Pressure from Section~\ref{sec:weather-case-study}) shows a quantile calibrated forecaster that is highly parity miscalibrated. 
(\textbf{top})
The expert forecasts $\hF_t$ are Gaussians, expressed in the plot as prediction intervals $[\hat{\mu}_t - 2\hat{\sigma}_t, \hat{\mu}_t + 2\hat{\sigma}_t]$. This prediction interval almost always contains $y_t$ and its reliability diagram in Figure~\ref{fig:pressure-gaussian-base} (plot titled ``Quantile Calibration") confirms that $\hF_t$ is in fact quantile calibrated when considering the full timeseries.
(\textbf{bottom}) 
For $t \in [0, 40]$  and $t \in [230, 300]$, the parity probabilities $\hp_t = \hF_t(y_{t-1})$ assign $\geq 0.8$ probability (red shaded areas) to $\indicator{y_t \leq y_{t-1}}$. But $y_t$ actually decreases with much lower frequency during these timesteps as can be seen from the top figure. The parity miscalibration when considering the full timeseries is confirmed by Figure~\ref{fig:pressure-gaussian-base} (plot titled ``Prehoc").
}
\vspace{-7mm}
\label{fig:bad_pressure_prehoc}
\end{figure}

\textbf{Synthetic example. }Let $\Ncal_{-}$ and $\Ncal_{+}$ denote the standard normal distributions truncated at $0$, with density functions $f_{-}(x) = \indicator{x < 0}\sqrt{2/\pi}e^{-x^2/2}$ and $f_{+}(x) = \indicator{x \geq 0}\sqrt{2/\pi}e^{-x^2/2}$ respectively. Let $F_{-}$ and $F_{+}$ be the cdfs of $\Ncal_{-}$ and $\Ncal_{+}$. Suppose the target sequence $(Y_t)_{t=1}^\infty$ is distributed as
\begin{align*}
    Y_{t} \sim \begin{cases}
    \Ncal_{-} & \text{if } t\text{ is odd, }  \\
    \Ncal_{+} & \text{if } t\text{ is even. } \\
    \end{cases}
\end{align*}
Consider the following predictive cdf targeting $Y_t$,
\begin{align*}
    \hat{F}_t &= \frac{1}{2} F_{-} + \frac{1}{2} F_{+} = \begin{cases}
        &\frac{1}{2}F_{-}(y) \text{, if } y<0, \\
        &0.5 + \frac{1}{2}F_{+}(y) \text{, if } y\geq 0.
    \end{cases}
\end{align*}
We note that when $y < 0, \frac{1}{2}F_{-}(y) \in [0, 0.5)$, and when $y \geq 0.5, 0.5 + \frac{1}{2}F_{+}(y) \in [0.5, 1]$.
It can be verified that the corresponding quantile function is 
\begin{align*}
    \hat{F}_t^{-1}(p) = \begin{cases}
        & F_{-}^{-1}(2p) \text{, if } p<0.5 \\
        & F_{+}^{-1}(2p-1) \text{, if } p\geq 0.5.
    \end{cases}
\end{align*}

We verify that $\hat{F}_t$ is quantile calibrated (following Eq.~\eqref{eq:probabilistic-calibration}).

\underline{When $t$ is odd},
$F_t = F_{-}$.
\begin{itemize}
    \item $\forall p \in [0, 0.5)$, $F_{t}(\hat{F}^{-1}_{t}(p)) = F_{-}(F_{-}^{-1}(2p)) = 2p$.
    \item $\forall p \in [0.5, 1]$, $\hat{F}^{-1}_t(p) = F_{+}^{-1}(2p-1) \geq 0$, thus $F_{t}(\hat{F}^{-1}_{t}(p)) = F_{-}(F_{+}^{-1}(2p-1)) = 1$.
\end{itemize}

\underline{When $t$ is even}, 
$F_t = F_{+}$.
\begin{itemize}
    \item $\forall p \in [0, 0.5)$, $\hat{F}^{-1}_t(p) = F_{-}^{-1}(2p) < 0$, thus $F_{t}(\hat{F}^{-1}_{t}(p)) = F_{+}(F_{-}^{-1}(2p)) = 0$.
    \item $\forall p \in [0.5, 1]$, ${F}_t(\hat{F}^{-1}_t(p)) = F_{+}(F_{+}^{-1}(2p-1)) = 2p - 1$.
\end{itemize}

Therefore, for $p \in [0, 0.5)$, 
$\frac{1}{T}\sum_{t=1}^{T}F_{t}(\hat{F}^{-1}_{t}(p)) = \frac{1}{T}\sum_{t \text{ is odd }}2p = p + o(\frac{1}{T}) \rightarrow p$, and the same can be verified for $p \in [0.5, 1]$, showing that \textit{$\hat{F}_t$ is quantile calibrated}. 

We can easily show that $\hF_t$ is also distribution and threshold calibrated. Since $\hF_t$ is constant for all $t$, following Eq.~\eqref{eq:distribution-calibration}, the space of predicted distributions is a singleton. Thus, measuring distribution calibration is equivalent to measuring quantile calibration, and \textit{$\hF_t$ is distribution calibrated}. Since distribution calibration implies threshold calibration~\citep{sahoo2021reliable}, \textit{$\hF_t$ is threshold calibrated}.

However, as we show next, $\hF_t$ is not parity calibrated. 

\underline{When $t$ is odd}, $Y_t \sim F_{-}$ and $Y_{t-1} \sim F_{+}$. Thus  $Y_t < Y_{t-1}$ whereas $\hp_t = \hat{F}_{t}(Y_{t-1}) \geq 0.5$.

\underline{When $t$ is even}, $Y_t \sim F_{+}$ and $Y_{t-1} \sim F_{-}$. Thus  $Y_t > Y_{t-1}$ whereas $\hp_t = \hat{F}_{t}(Y_{t-1}) < 0.5$.

Therefore, 
$\forall \hat{p}_t \geq 0.5$, $\indicator{y_t \leq y_{t-1}} = 1$
and $\forall \hp_t < 0.5$, $\indicator{y_t \leq y_{t-1}} = 0$,
thus \textit{$\hF_t$ is parity miscalibrated} for all $\hp_t \in (0, 1)$, i.e. all $\hp_t \neq 0 \text{ or } 1$. \qedwhite

Intuitively, the sequential aspect of predictions and observations is central to the notion of parity calibration, whereas traditional notions of calibration effectively
treat the datapoints as an i.i.d. or exchangeable batch of points.
Figure~\ref{fig:bad_pressure_prehoc} provides a visualization of how this pitfall can 
be manifested in a practical example.

The implication is that methods designed to achieve traditional notions of calibration in regression cannot be expected to provide parity calibration. The following section introduces the posthoc binary calibration framework that can instead be used to achieve parity calibrated forecasts. 

\section{Parity calibration via binary calibration}
\label{methods}
Define the \textit{parity outcomes} as 
\begin{align} \label{eq:parity-outcome}
    \text{ for $ t \geq 2$, }\ \widetilde{y}_t := \indicator{y_t \leq y_{t-1}},
\end{align}
and observe that the parity calibration condition (Eq.~\eqref{eq:parity-calibration-defn}) is equivalently written as, 
\begin{equation}
    \frac{\sum_{t=2}^{T} \widetilde{y}_t \indicator{\hp_t=p}}{\sum_{t=2}^{T} \indicator{\hp_t=p}} \rightarrow p, \forall p \in [0, 1]. \label{eq:parity-calibration-equivalent}
\end{equation} 
Thus parity calibration is in fact targeting the binary sequence $\widetilde{y_t}$, instead of $y_t$. In this section, we show how this connection allows us to leverage powerful techniques from the rich literature of binary calibration that goes back four decades \citep{degroot1981assessing, dawid1982well, foster1998asymptotic}.
Of specific interest to us will be a class of methods that have been proposed for \textit{posthoc calibration} of machine learning (ML) classifiers, which we review next. 

\subsection{Posthoc binary calibration}

Let $f : \X \to [0,1]$ be a binary classifier that takes as input a feature vector in feature space $\X$ and outputs a score in $[0,1]$. Suppose a feature-label pair $(X, Y)$ is drawn from some distribution $P$ over $\Xcal \times \{0, 1\}$. Then, $f$ is said to be calibrated (in the binary sense) if 
\begin{equation}
    P(Y = 1 \mid f(X)) = f(X).\label{eq:perfect-calibration}
\end{equation}
The terms on either side of the equal sign are random variables and the equality is understood almost-surely. The connection between \eqref{eq:parity-calibration-equivalent} and \eqref{eq:perfect-calibration} is evident: $\hp_t$ is like $f(X)$, conditioning on the random variable $f(X)$ is akin to using indicators in the numerator/denominator, and $\widetilde{y}_t$ is like $Y$. 

We do not expect ML models to be calibrated ``out-of-the-box''. So, if $f$ is a logistic regression or neural network trained on some training data, it is unlikely to satisfy an approximate version of  \eqref{eq:perfect-calibration} on unseen data. Posthoc calibration techniques transform $f$ to a function that is better calibrated by using a so-called \textit{calibration dataset}  $\mathcal{D}_{\text{cal}} = \{(\x_1, y_1), (\x_2, y_2), \ldots, (\x_c, y_c)\}$. $\mathcal{D}_{\text{cal}}$ is a set of points on which $f$ was not trained---in practice $\Dcal_{\text{cal}}$ is often just the validation dataset. $\mathcal{D}_{\text{cal}}$ is used to a learn a mapping $m: [0,1] \to [0,1]$ so that $m\circ f$ is better calibrated than $f$. By way of an example, we now introduce the popular Platt scaling technique~\citep{platt1999probabilistic} that will be central to this paper (henceforth, Platt scaling is referred to as PS). Given a pair of real numbers $(a, b) \in \mathbb{R}^2$, the PS mapping $m^{a,b} : [0, 1] \to [0,1]$ is defined as,
\begin{equation*}
    m^{a,b}(z) = \text{sigmoid}(a\cdot \text{logit}(z) + b). 
\end{equation*}
Here $\text{logit}(z) = \log(\frac{z}{1-z})$ and $\text{sigmoid}(z) = 1/(1+e^{-z})$ are inverses of each other. Thus PS is a logistic model on top of the $f$-induced one-dimensional feature $\text{logit}(f(x)) \in [0,1]$, instead of on the raw feature $x \in \Xcal$. 
In the posthoc setting, $(a, b)$ are set to the values that minimize log-loss (equivalently cross entropy loss) on $\mathcal{D}_{\text{cal}}$:
\begin{equation}
    (\widehat{a}, \widehat{b}) = \argmin_{(a, b) \in \mathbb{R}^2} \sum_{ (\x_s, y_s) \in \mathcal{D}_{\text{cal}}} l(m^{a,b}(f(\x_s)), y_s),  \label{eq:optimal-ops-t}
\end{equation}
where $l(p, y) = -y\log p-(1-y)\log(1-p)$. 

We briefly note some other popular posthoc calibration methods. These broadly fall under two categories: parametric scaling methods such as beta scaling \citep{kull2017beyond}, temperature scaling \citep{guo2017calibration}, and PS \citep{platt1999probabilistic}; 
and nonparametric methods such as binning \citep{zadrozny2001obtaining, gupta2020distribution, gupta2021distribution}, isotonic regression \citep{zadrozny2002transforming}, and Bayesian binning \citep{naeini2015obtaining}. 

\subsection{Parity calibration using online versions of Platt Scaling (PS)}
\label{sec:main-parity-calibration-methodology}
To achieve parity calibration using posthoc techniques, we start with a base cdf predictor $G : \X \to \Delta(\Y)$ derived from an expert---such as an epidemiologist, a weather forecaster, or a stock trader. 
Here, $\Delta(\Y)$ refers to the space of distributions over $\Y$.
If the expert is an ML engineer, such a $G$ can be obtained using Gaussian processes~\citep{rasmussen2003gaussian} or probabilistic neural networks~\citep{nix1994estimating, lakshminarayanan2017simple}, among other methods. 
The test-stream occurs after $G$ has been trained and fixed. 
This $G$ gives us a $\hF_t$ as described in the introduction: $\hF_t = G(\x_t)$. Recall that the strategy Eq.~\eqref{eq:naive-strategy} is to forecast $\hat{p}_t = \hF_t(y_{t-1})$. 
If $\hF_t$ were the true cdf of $y_{t}$ given the past, the above $\hat{p}_t$ would be the true probability of $\widetilde{y}_t = 1$, and thus the most useful parity forecast possible. 

However, in Section~\ref{sec:counter-examples} we showed that we must modify $\hp_t$ in order to achieve parity calibration. We propose using PS to perform this modification (any posthoc calibration method can be used; we focus on PS in this paper). A natural possibility would be to use an initial part of the test-stream to learn fixed PS parameters once, as described in the previous subsection. However, real-world regression sequences (weather, stocks, etc) have non-stationary shifting behavior across time. 
Therefore, a fixed model is unlikely to remain calibrated over time. 

In Algorithm~\ref{alg:windowed_platt_scaling} we outline three ways to mitigate this. Increasing Window (IW) updates the PS parameters using all datapoints until some recent time step, such as every 100 timesteps ($t = 100, 200,$ etc). A related alternative, Moving Window (MW) is to use only the most recent datapoints when updating the PS parameters (instead of all the points). 
The third alternative is Online Platt Scaling (OPS) based on our own recent work \citep{gupta2023online}.

\begin{algorithm}[tb]
   \caption{Platt scaling (PS) variants for parity calibration}
   \label{alg:windowed_platt_scaling}
\begin{algorithmic}[1]
   \STATE {\bfseries Input}: Any base forecaster $G : \Xcal \to \Delta(\Ycal)$, covariate-outcome pairs $(\x_1, y_1), (\x_2, y_2), \ldots \in \Xcal \times \Ycal$, update-frequency \texttt{uf}, moving-window-size \texttt{ws}.
  \STATE {\bfseries Output}: PS forecasts $(\hp_t^\iw$, $\hp_t^\mw$, $\hp_t^\ops)_{t=2}^\infty$
   \STATE Initialize IW, MW, OPS parameters: \\
    $\  (a^{\text{IW}}, b^{\text{IW}}) = (a^{\text{MW}}, b^{\text{MW}}) = (a^{\text{OPS}}, b^{\text{OPS}}) \gets (1, 0)$\;
    
    \FOR{$t=2$ {\bfseries to} $T$}
    \STATE $\widetilde{y}_t = \indicator{y_t \leq y_{t-1}}$
    \STATE $\hat{p}_t = G(\x_t)[y_{t-1}]$
    \STATE $\hp_t^{\text{IW}} \gets \sigmoid(a^\text{IW}\cdot  \logit(\hp_t) + b^\text{IW})$\;
    \STATE $\hp_t^{\text{MW}} \gets \sigmoid(a^\text{MW}\cdot  \logit(\hp_t) + b^\text{MW})$\;
    \STATE $\hp_t^{\text{OPS}} \gets \sigmoid(a^\text{OPS}\cdot  \logit(\hp_t) + b^\text{OPS})$\;
    \IF{$t$ is a multiple of \texttt{uf}} 
    \STATE $(a^{\text{IW}}, b^{\text{IW}}) \gets $ optimal PS parameters \\ \ \ \ \ based on \eqref{eq:optimal-ops-t} setting $\Dcal_{\text{cal}} = (\x_s, \widetilde{y}_s)_{s = 1}^t$\;
    \STATE $(a^{\text{MW}}, b^{\text{MW}}) \gets $ optimal PS parameters \\ \ \ \ \ based on \eqref{eq:optimal-ops-t} setting $\Dcal_{\text{cal}} = (\x_s, \widetilde{y}_s)_{s = t-\texttt{ws}+1}^t$\;
    \ENDIF
      \STATE $(a^\ops, b^\ops) \gets \text{OPS}((\x_1, \widetilde{y}_1),\ldots,(\x_{t}, \widetilde{y}_{t}))$
     \STATE (OPS is Algorithm~\ref{alg:ops-ons} in Appendix~\ref{app:ops-algorithm})
    \ENDFOR
\end{algorithmic}
\end{algorithm}

In the following section, we compare these online versions of Platt scaling on three real-world sequential prediction tasks. We find that OPS performs better than the base model, MW, and IW, across multiple settings. 
Further, while MW and IW involve re-fitting the PS parameters from scratch, OPS makes a constant time update at each step, hence the overall computational complexity of OPS is $O(T)$. 

\textbf{Brief note on theory and limitations of OPS. }OPS satisfies a regret bound with respect to the Platt scaling class for log-loss \citep[Theorem 2.1]{gupta2023online}. This means that the OPS forecasts do as well as forecasts of the single best Platt scaling model in hindsight. 
However, we note that OPS could fail if the best Platt scaling model is itself not good. This limitation can be overcome by combining OPS with a method called calibeating, as discussed in 
\citet{gupta2023online}. We do not pursue calibeating in this paper since OPS already performs well on the data we considered.

\section{Real-world case studies}\label{sec:experiments}
We study parity calibration in three real-world scenarios: 1) forecasting COVID-19 cases in the United States, 2) forecasting weather, and 3) predicting plasma state evolution in nuclear fusion experiments.
This diverse set of domains, datasets, and expert forecasters 
provides an attractive test-bed to 
demonstrate the parity calibration concept and the performance of the calibration methods from Section~\ref{sec:main-parity-calibration-methodology}.

In each setting, the prediction target is real-valued, and we assume an expert forecaster provides regression forecasts $\hF_t$ for the target. 
We also refer to $\hF_t : \Ycal \to [0,1]$ as the \emph{base regression model}.
The expert forecaster implicitly provides parity probabilities $\hp_t$ (following Eq.~\eqref{eq:naive-strategy}). We refer to $\hp_t$ as the \emph{prehoc} probabilities, in contrast to the \emph{posthoc} probabilities that the calibration methods produce.
We calibrate $\hp_t$ with the calibration methods from Section~\ref{sec:main-parity-calibration-methodology} to produce the posthoc probabilities $\hp'_t$.
Each calibration method requires a set of hyperparameters, 
which we tune with a validation set. Details regarding hyperparameter tuning are provided in Appendix~\ref{app:hyperparameters}.
{\renewcommand\thefootnote{}\footnote{\smash{Code is available at} {\fontsize{7}{9.6} \url{https://github.com/YoungseogChung/parity-calibration}}}}

\textbf{Metrics. }
Given a test dataset $\mathcal{D}_{\text{test}} = \{\x_t, y_t\}_{t=1}^{T}$, 
we initially assess the quantile calibration of $\hF_t$ and the parity calibration of $\hp_t$ and $\hp'_t$ by visualizing the reliability diagrams and measuring calibration errors. 

To assess quantile calibration of $\hF_t$, we produce the reliability diagram using the Uncertainty Toolbox~\citep{chung2021uncertainty}, which takes a finite set of quantile levels $\mathcal{P} = \{p_i \in [0, 1]\}$, computes the empirical coverage of the predictive quantile $\hF^{-1}_t(p_i)$ as $p_{i, \text{obs}} = \frac{1}{T}\sum_{t=1}^{T}\indicator{y_t \leq \hF^{-1}_t(p_i)}$, and plots each $p_i$ against $p_{i, \text{obs}}$. Calibration error is then summarized into a single scalar with Quantile Calibration Error (QCE), which is computed as  $\frac{1}{\mid \mathcal{P} \mid} \sum_{i} \mid p_{i, \text{obs}} - p_i\mid$. In our experiments, we set $\mathcal{P}$ to be $100$ equi-spaced quantile levels in $[0, 1]$.

To assess parity calibration of a parity probability $\hp_t$, we follow the standard method of producing reliability diagrams in binary calibration~\citep{degroot1981assessing, niculescu2005predicting}. Noting that $\hp_t$ is a predicted probability of the binary parity outcome 
$\widetilde{y}_t := \indicator{y_t \leq y_{t-1}}$, we first bin $\hp_t$ into a finite set of fixed width bins $\mathcal{B} = \{B_m\}$, then for each bin $B_m$, we compute the average outcome as $\text{obs}(B_m) = \frac{1}{\mid B_m \mid} \sum_{t: \hp_t \in B_m} \indicator{\widetilde{y}_t = 1}$ and the average prediction as $\text{pred}(B_m) = \frac{1}{\mid B_m \mid} \sum_{t: \hp_t \in B_m} \hp_t$, and finally, we plot $\text{pred}(B_m)$ against $\text{obs}(B_m)$ to produce the reliability diagram.
Parity Calibration Error (PCE) summarizes the diagram following the standard definition of ($\ell_1$-)expected calibration error (ECE): $\sum_m \frac{\mid B_m \mid}{T}\mid \text{obs}(B_m) - \text{pred}(B_m) \mid$. In our experiments, we set $\mathcal{B}$ to be $30$ fixed-width bins: $[0, \frac{1}{30}), [\frac{1}{30}, \frac{2}{30}), \dots [\frac{29}{30}, 1]$.

For the parity probabilities $\hp_t$ and $\hp'_t$, we additionally 
report sharpness and two metrics for accuracy: {binary accuracy} and {area under the ROC curve}.
Sharpness (Sharp) is computed as $\sum_{m} \frac{\mid B_m \mid}{T}\cdot \text{obs}(B_m)^{2}$ and measures the degree to which the forecaster can discriminate events with different outcomes~\citep{brocker2009reliability}. {Binary accuracy} (Acc) and {area under the ROC curve} (AUROC) are computed following their standard definitions in binary classification.
Appendix~\ref{app:metrics} provides the full set of details on how each metric is computed.
Lastly, in reporting the metrics in numeric tables, we denote each metric with their orientation, e.g. $\uparrow$ indicates that a higher value is more desirable and vice versa.

\begin{figure}[t]
    \centering
    \begin{subcaptionblock}{\columnwidth}
        \centering
        \includegraphics[width=\linewidth]{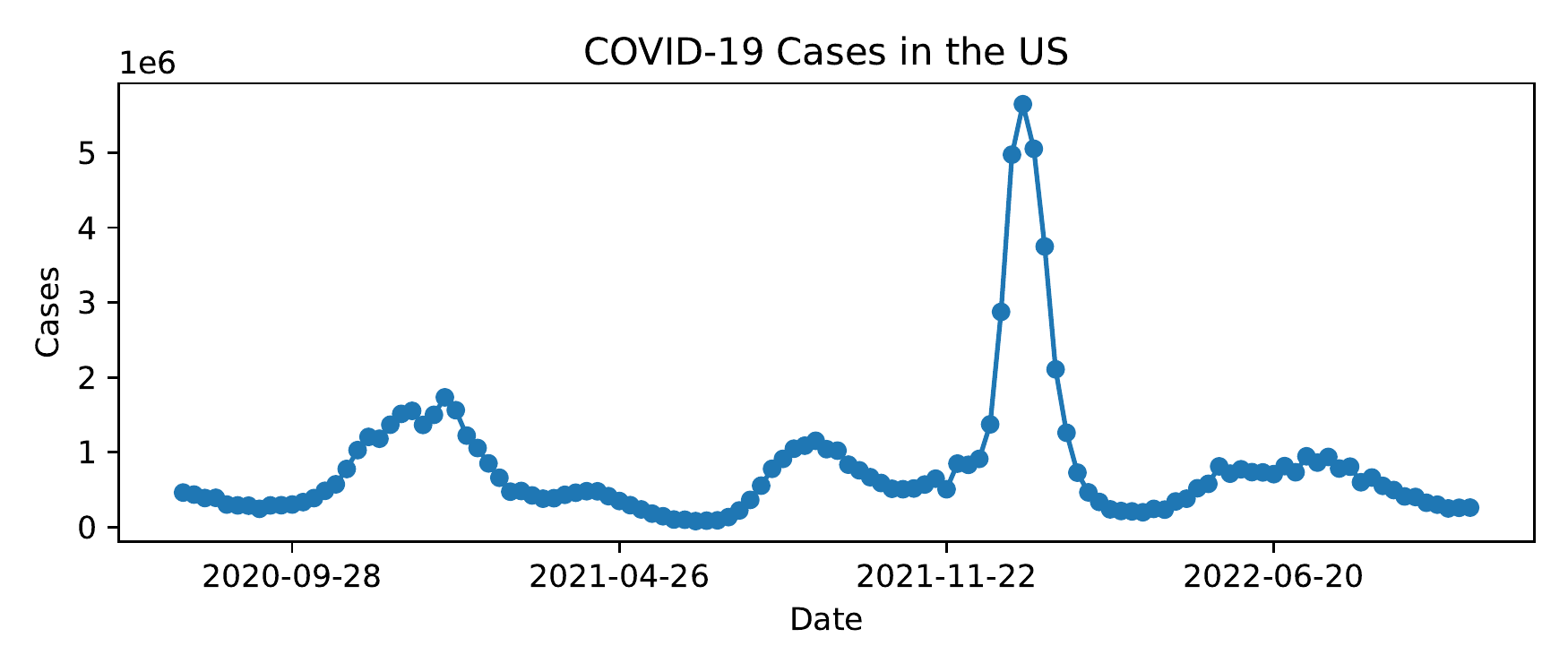}
        \caption{Total COVID-19 cases in the US displays high non-stationarity.}
        \label{fig:covid-cases-raw}
    \end{subcaptionblock}
    \begin{subcaptionblock}{\columnwidth}
        \centering
        \includegraphics[width=\linewidth]{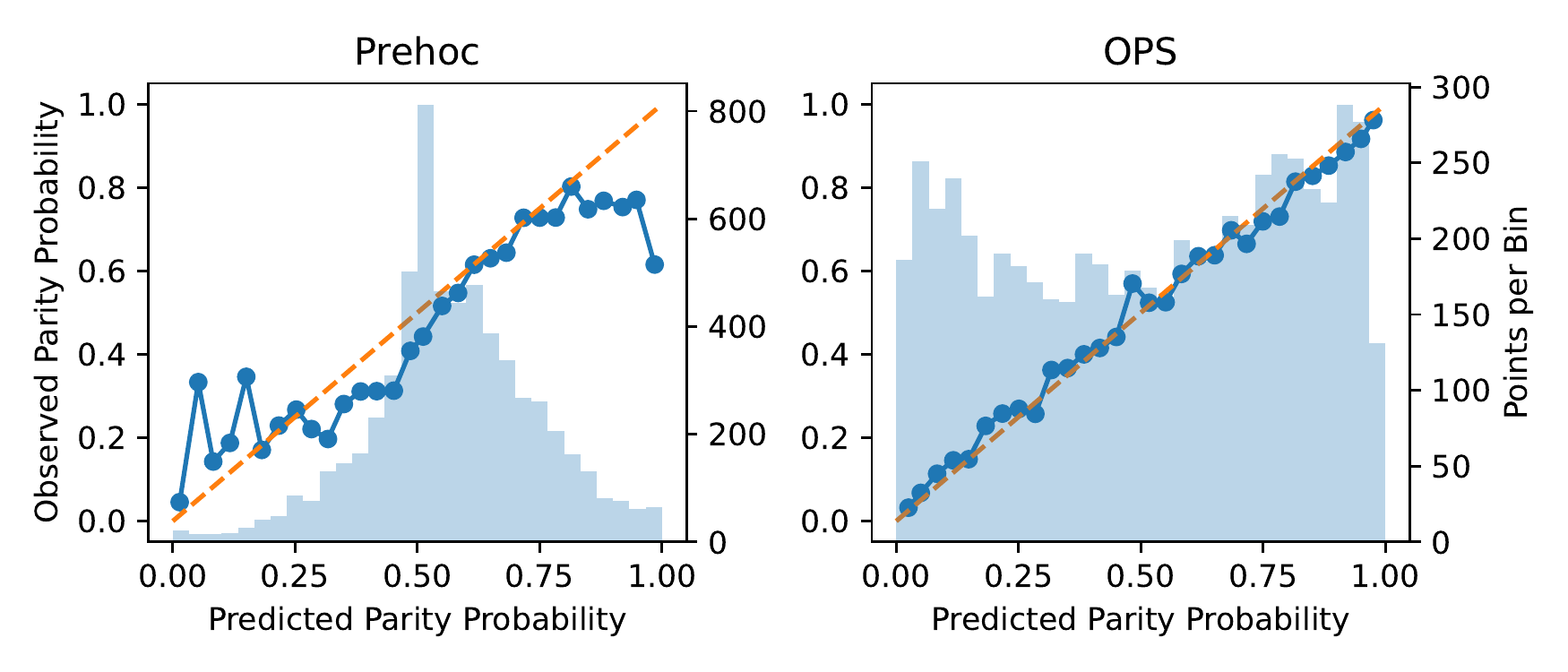}
        \caption{Reliability diagrams for the prehoc parity probabilities from the expert forecasts (\textbf{left}) and OPS calibrated probabilities (\textbf{right}). \textbf{Blue bars} denote the frequency of predictions in each bin.}
        \label{fig:covid-parity-calibration}
    \end{subcaptionblock}
\caption{The prehoc parity probabilities for the COVID-19 single-timeseries setting are miscalibrated and un-sharp. Posthoc calibration via OPS improves both aspects.}
\label{fig:covid-us}
\vspace{-2mm}
\end{figure}

\subsection{Case Study 1: COVID-19 cases in the US}
\label{sec:covid-case-study}

In response to the COVID-19 pandemic, research groups across the world have created models to predict the short-term future of the pandemic. The COVID-19 Forecast Hub \citep{cramer2021hubdataset} solicits and collects quantile forecasts of weekly incident COVID-19 cases in each US state (plus Washington D.C.), among other targets. Each week, the Hub generates an ensemble forecast from the dozens of submitted forecasts.
This ensemble has proven to be more reliable and accurate than any constituent individual forecast in predicting other targets of interest (e.g. mortality~\citep{cramer2022ensemble}). Thus, we take the ensemble forecast as the expert forecast and use its historical forecasts made between 2020-07-20 and 2022-10-24, which span a total of 119 weeks.
Denoting the target $y$ as the number of cases, there are effectively 51 timeseries, $\{y_{s, t}\}$: one for each US state $s \in$ \{Alabama, Alaska, Arizona, ..., Wisconsin, Wyoming\}, and $t\in\{1, \dots, 119\}$.
For any given $s, t$, the expert forecast is provided by the Hub as seven forecasted quantiles for the distribution of $y_{s, t}$. Therefore, we must interpolate the quantiles to produce $\hF_t$ (see Appendix~\ref{app:covid-appendix} for details). 

The observed targets $y_{s, t}$ are the incident number of cases actually reported
from each state, for each week.
Figure~\ref{fig:covid-cases-raw} visualizes a summary of the target timeseries: the total incident number of cases in the US $(=\sum_s y_{s, t})$. We can observe high non-stationarity, with periods of rapid increases and falls, and other periods of long monotonic trends.

\subsubsection{Parity calibration of expert forecasts and OPS}\label{sec:covid-single-timeseries}
Note that the underlying timeseries $\{y_{s, t}\}$ is indexed by both state and time. We transform this to a fully sequential timeseries by concatenating $\{y_{s, t}\}$ chronologically across $t$ and in alphabetical order across $s$. 
In other words, within a given week, we observe the number of cases for the states in alphabetical order. 
We refer to this experiment setting as the \textit{single-timeseries} setting.

The reliability diagram in Figure~\ref{fig:covid-parity-calibration} (left) shows that the prehoc probabilities implied by the expert forecast ($\hp_t$) are parity calibrated in the $[0.25, 0.75]$ region (i.e. higher predicted probabilities result in higher empirical frequencies), but are miscalibrated otherwise. The distribution of $\hp_t$ displayed by the blue bars further indicate that $\hp_t$ is centered around $0.5$, an uninformative or less sharp prediction. 

\begin{table}[t]
\centering
\begin{tabular}{lccc} \toprule
        & Prehoc    & OPS\textsubscript{alpha-order}    & OPS\textsubscript{rand100}  \\\midrule
PCE $\downarrow$   & $0.0599$   & $0.0216$  & $0.0246 \pm 0.0002$  \\
Sharp $\uparrow$  & $0.2953$   & $0.3087$  & $0.3090 \pm 0.00002$  \\
Acc $\uparrow$    & $0.6309$   & $0.6727$ & $0.6737 \pm 0.0001$  \\
AUROC $\uparrow$  & $0.6922$   & $0.7355$ & $0.7357 \pm 0.00002$  \\\bottomrule
\end{tabular}
\caption{In the COVID-19 single-timeseries setting, OPS improves the prehoc parity probabilities w.r.t all metrics. $\pm$ indicates mean $\pm$ 1 standard error across 100 state orders.}
\label{tab:covid-single-timeseries}
\end{table}

\begin{table}[t]
\centering
\begin{tabular}{lcccc} \toprule
          & Prehoc    & MW         & IW        & OPS      \\ \midrule
PCE  $\downarrow$  & 0.0599 & 0.0748  & 0.0406  & \textbf{0.0328}   \\
Sharp $\uparrow$   & 0.2953 & 0.2882  & 0.2839  & \textbf{0.2993}   \\
Acc  $\uparrow$    & 0.6309 & 0.6237  & 0.6055  & \textbf{0.6522}   \\
AUROC $\uparrow$   & 0.6922 & 0.6622  & 0.6403  & \textbf{0.7035}   \\ \bottomrule
\end{tabular}
\caption{In the COVID-19 sequential-batch setting, OPS outperforms prehoc and alternative PS methods. Best value for each metric is in bold.}
\label{tab:covid-method-comparison}
\end{table}

Figure~\ref{fig:covid-parity-calibration} (right) displays the reliability diagram of $\hp_t^{\ops}$. We observe significant improvements in both parity calibration and sharpness, i.e. $\hp_t^{\ops}$ is much more dispersed compared to $\hp_t$.
The second column of Table~\ref{tab:covid-single-timeseries} (labeled OPS\textsubscript{alpha-order}) show these improvements via the PCE and Sharp metrics, and we can also observe improvement in accuracy.

One may question whether this improvement by OPS is specific to the alphabetical order of states.
In the third column of Table~\ref{tab:covid-single-timeseries} (labeled OPS\textsubscript{rand100}), we show the mean and standard error of each of the metrics across 100 different random orders of the states, and observe that 
the improvements provided by OPS over prehoc are fairly robust.

\begin{figure}[t]
\begin{center}
\centerline{\includegraphics[width=\columnwidth]{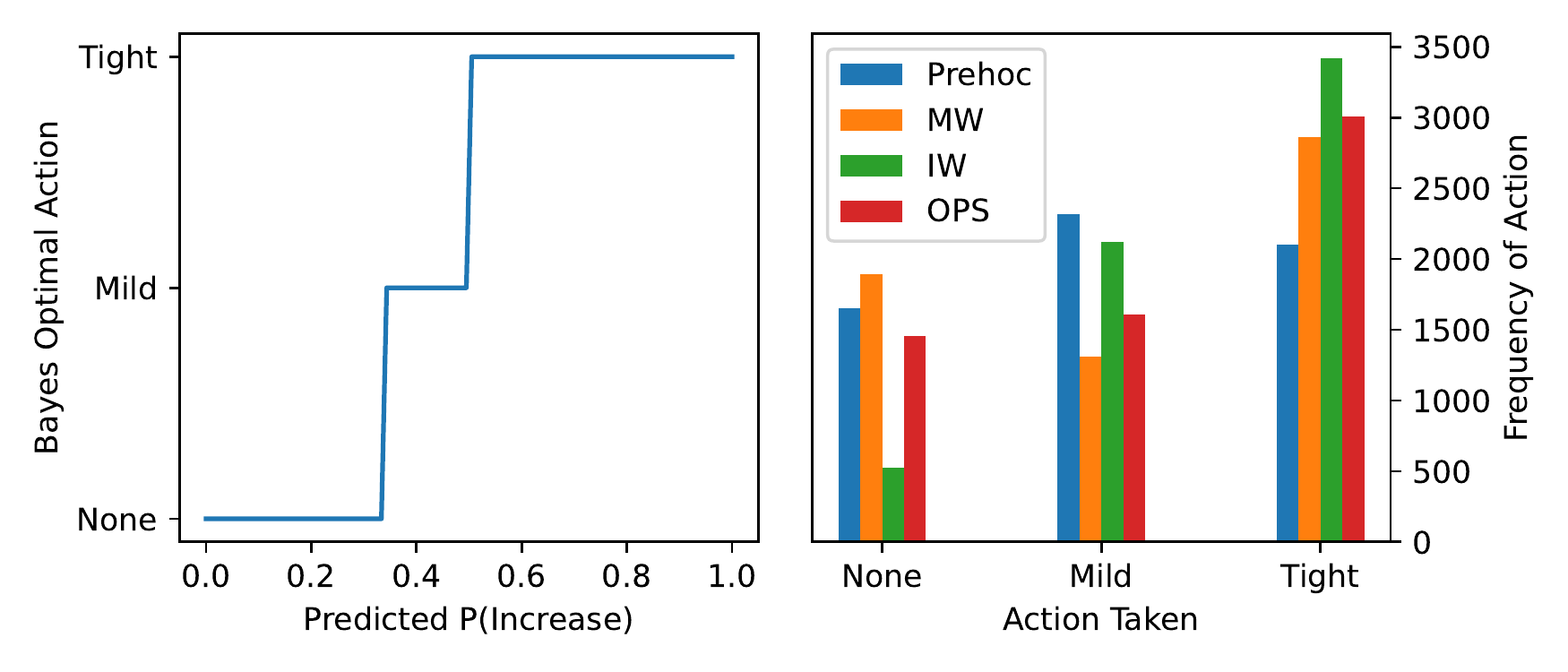}}
\caption{(Decision making on the COVID-19 dataset) (\textbf{left}) The Bayes optimal action for each predicted probability of increase in number of cases. (\textbf{right}) Frequency of each action taken by each method.}
\label{fig:covid-decision-making}
\end{center}
\vspace{-8mm}
\end{figure}

\subsubsection{Comparing calibration methods}\label{sec:covid-methods-comparison}
We perform an additional experiment to compare the performance of MW, IW and OPS.
In this experiment, we assume a more realistic test setting for the data-stream. At each timestep $t$, we assume we observe cases from all 51 states, $\{y_{s, t}\}_{s=1}^{51}$, and update the
PS parameters with this batch of data.
We then fix the PS parameters and calibrate the next full batch of predictions for timestep $t+1$.
This settings assumes that PS parameters are updated once per week based on all the data observed during the week.
We refer to this experiment setting as the \textit{sequential-batch} setting.

The first 20 weeks of data (i.e. 20 weeks  $\times$ 51 states = 1020 datapoints) were used to tune the hyperparameters of each method.
The subsequent 99 weeks of data was used for testing. Table~\ref{tab:covid-method-comparison} displays the results of the sequential batch setting (note that the prehoc values are the same for this setting as in Table~\ref{tab:covid-single-timeseries}).
OPS is the best performing method on all metrics when compared with MW, IW, and prehoc.

\begin{figure*}[t]
\centering 
\includegraphics[width=0.93\linewidth]{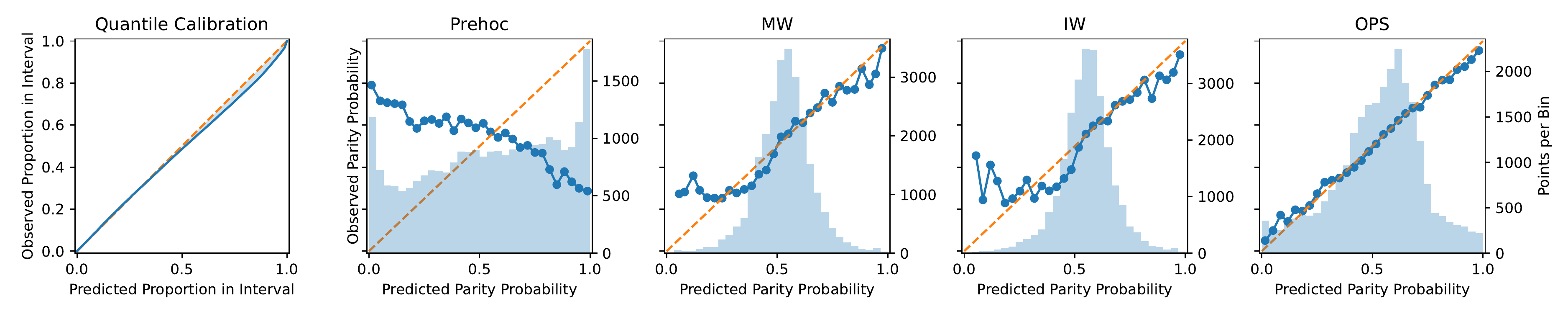}
\caption{OPS significantly improves both parity calibration and sharpness of the base regression model predicting Pressure. The left two plots display the quantile calibration and parity calibration of the base model (Prehoc): it is nearly perfectly quantile calibrated, but terribly parity calibrated. \textbf{Blue bars} denote the frequency of predictions in each bin.
}
\label{fig:pressure-gaussian-base}
\vspace{-2mm}
\end{figure*}

\subsubsection{Decision-making with parity probabilities} \label{sec:covid-decision-making}
In this section, we demonstrate the utility of OPS in a decision-making setting where 
parity outcomes (Eq.~\eqref{eq:parity-outcome}) 
dictate the loss incurred.
Using the same COVID-19 dataset,
we assume a setting where a policymaker (i.e. the decision-maker) at each timestep must decide among three levels of restrictions for disease spread prevention: Tight, Mild, or None.
For any chosen level of restriction, the loss is dictated by the parity outcome in the number of cases, and the policymaker's goal is to minimize cumulative loss.
A \textit{Bayes optimal} policymaker will always choose an action which minimizes the expected loss, calculated with a predictive distribution over the loss~\citep{lehmann2006theory}.
Hence the policymaker will assess the optimality of each action based on predicted parity probabilities.

We design an exemplar loss function $l_{\text{truth, decision}}$ as follows: 

\begin{table}[ht]
\vspace{-1.5mm}
\centering
{\setlength{\tabcolsep}{5.5pt}
\begin{tabular}{cccc}
\toprule
\# Cases  & Tight = 1        & Mild = 2       & None = 3      \\\midrule
Increase = 1   & $l_{1, 1} = 0.3$   & $l_{1, 2}=0.6$  & $l_{1, 3} = 1$ (max)   \\ \midrule
Decrease = 2   & $l_{2, 1}=0.5$   & $l_{2, 2}=0.2$  & $l_{2, 3} = 0$ (min)   \\ \bottomrule
\end{tabular}}
\vspace{-2mm}
\end{table}

Given this loss function, the Bayes optimal action is visualized in Figure~\ref{fig:covid-decision-making} (left). On computing the the cumulative loss incurred with the predicted parity probabilities, we find that OPS incurs the lowest cumulative loss.
\begin{table}[ht]
\vspace{-1mm}
\centering
\begin{tabular}{ccccc}
\toprule
          & Prehoc         & MW        & IW  &OPS      \\ \midrule
Loss $\downarrow$   & 2119   & 2177  & 2196 & \textbf{2050}  \\ \bottomrule
\end{tabular}
\vspace{-2mm}
\end{table}\\
Figure~\ref{fig:covid-decision-making} (right)
shows the frequency of each action chosen by each method. We observe that 
OPS chooses Mild with relatively low frequency, which is a result of sharper and more accurate parity probabilities. 
We further note that IW results in a worse loss than prehoc despite being better parity calibrated (Table~\ref{tab:covid-method-comparison}). To understand this, notice that IW is also less sharp and less accurate than Prehoc. Thus calibration, while a desirable quality, is not the only aspect to assess for good uncertainty quantification---sharpness and accuracy could also affect decision making. 

\subsection{Case Study 2: Weather forecasting}\label{sec:weather-case-study}
\begin{figure}[ht]
\begin{center}
\centerline{\includegraphics[width=\columnwidth]{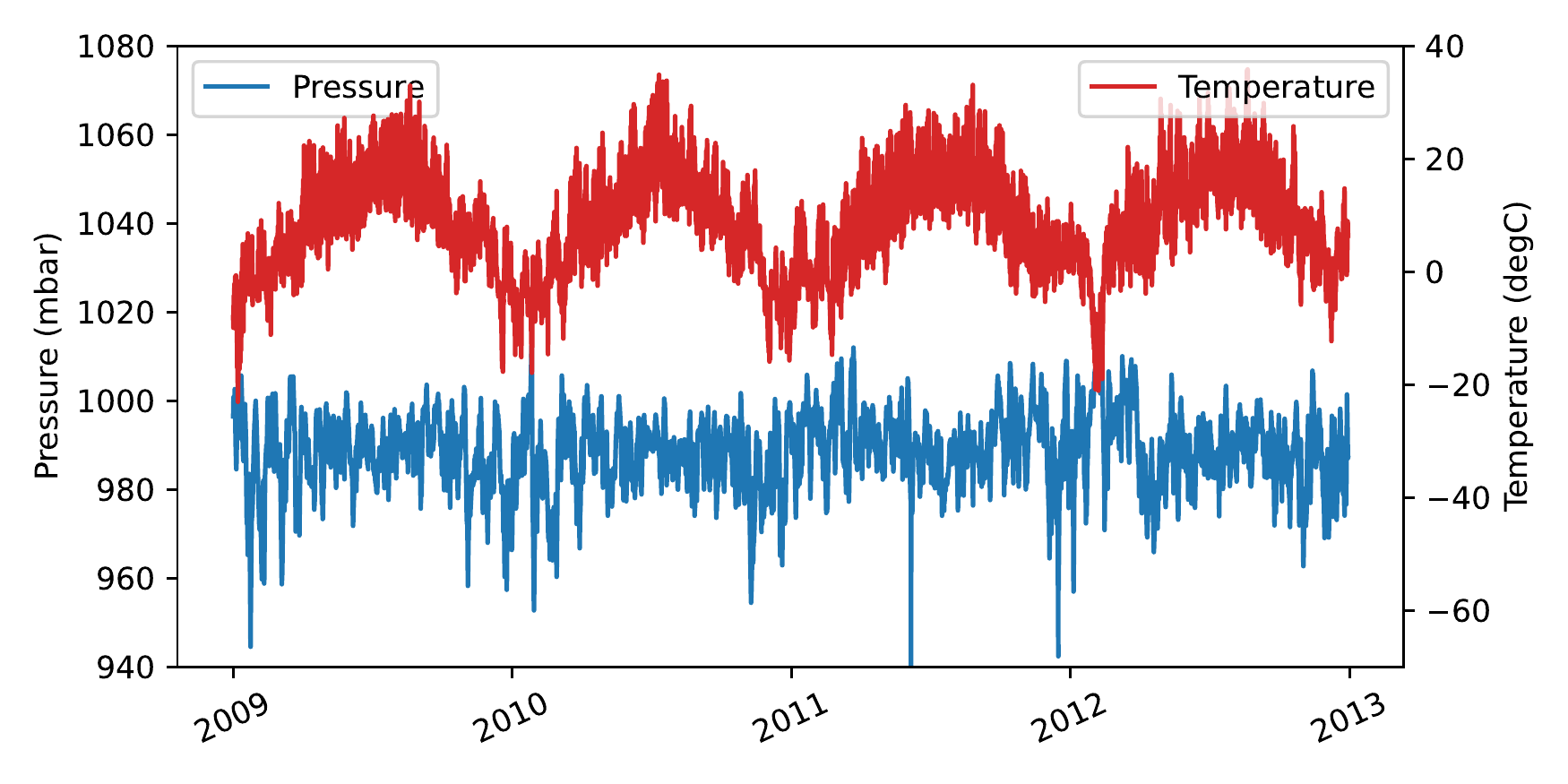}}
\caption{Snapshots of 4 years from the Temperature and Pressure timeseries display noise around a cyclical trend.}
\label{figure:weather-timeseries}
\end{center}
\vspace{-7mm}
\end{figure}

Our second case study examines weather forecasting using the benchmark Jena climate modeling dataset~\citeyearpar{jenaWeather}, which records the weather conditions in Jena, Germany, with 14 different measurements, in 10 minute intervals, for the years  2009---2016.
We did not have access to  historical predictions from an expert weather forecaster,
so instead we trained our own base regression model.

We follow the Keras tutorial on \textit{Timeseries Forecasting for Weather Prediction}\footnote[1]{\fontsize{7}{2}\url{https://keras.io/examples/timeseries/timeseries_weather_forecasting/}} to define our specific problem setup and train our base regression model.
In summary, the regression model is implemented with an LSTM network~\citep{hochreiter1997long} which predicts the mean and variance of a Gaussian distribution.
We trained 7 different models that each predict one of 7 weather features: Pressure, Temperature, Saturation vapor pressure, Vapor pressure deficit, Specific humidity, Airtight, and Wind speed.
Appendix~\ref{app:weather-experiment-details} provides more details on the problem setup.

Lastly, we note that unlike the COVID-19 data, the weather data (Figure~\ref{figure:weather-timeseries})
displays high levels of noise around a cyclical, repeating trend.

\begin{table*}[ht]
\centering
{\setlength{\tabcolsep}{5pt}
    \begin{tabular}{lccccc}
    \toprule
            & QCE $\downarrow$   & PCE $\downarrow$   & Sharp $\uparrow$   & Acc $\uparrow$   & AUROC $\uparrow$\\ \midrule
    Prehoc & \textbf{0.0181$\pm$0.0026} & 0.3493$\pm$0.0015 & 0.3019$\pm$0.0004 & 0.4044$\pm$0.0006 & 0.3525$\pm$0.0012 \\
    MW     & N/A             & 0.0278$\pm$0.0005 & 0.3005$\pm$0.0004 & 0.6124$\pm$0.0008 & 0.6410$\pm$0.0012 \\
    IW     & N/A             & 0.0322$\pm$0.0005 & 0.3013$\pm$0.0004 & 0.6147$\pm$0.0009 & 0.6450$\pm$0.0013 \\
    OPS    & N/A             & \textbf{0.0148$\pm$0.0002} & \textbf{0.3172$\pm$0.0004} & \textbf{0.6525$\pm$0.0007} & \textbf{0.7056$\pm$0.0010} \\
    \bottomrule
    \end{tabular}
}
\caption{
OPS improves the overall quality of parity probabilities from the base regression model predicting Pressure. $\pm$ indicates mean $\pm$ 1 standard error, across 50 test trials. Best value for each metric is in bold.
}
\label{tab:pressure_numerical}
\end{table*}

\begin{table*}[ht]
\centering
{\setlength{\tabcolsep}{5pt}
    \begin{tabular}{lccccc}
    \toprule
            & PCE $\downarrow$ & Sharp $\uparrow$ & Acc $\uparrow$  & AUROC $\uparrow$ \\ \midrule
    Prehoc  & 0.0258$\pm$0.0005 & 0.3008$\pm$0.0007 & 0.6069$\pm$0.0011 & 0.6474$\pm$0.0016 \\
    MW      & 0.0201$\pm$0.0005 & 0.3002$\pm$0.0007 & 0.6050$\pm$0.0012 & 0.6439$\pm$0.0017 \\
    IW      & 0.0166$\pm$0.0003 & 0.3003$\pm$0.0008 & 0.6068$\pm$0.0010 & 0.6456$\pm$0.0016 \\
    OPS     & \textbf{0.0150$\pm$0.0001} & \textbf{0.3232$\pm$0.0006} & \textbf{0.6665$\pm$0.0007} & \textbf{0.7275$\pm$0.0007} \\
    \bottomrule
    \end{tabular}
}
\caption{While MW, IW, OPS all improve parity calibration of the base classification model for Pressure (Prehoc), OPS is the only method that improves all metrics simultaneously. $\pm$ indicates mean $\pm$ 1 standard error, across 50 test trials. Best value for each metric is in bold.}
\label{tab:pressure-binary-prehoc-ops}
\vspace{-3mm}
\end{table*}

\textbf{Results on Pressure timeseries.}
We first examine results from one of the 7 models predicting Pressure.
Figure~\ref{fig:pressure-gaussian-base} displays quantile calibration (i.e. probabilistic calibration) of the base model, and parity calibration before and after MW, IW and OPS are applied to the prehoc parity probabilities.
We first note that the base model is almost perfectly quantile calibrated, but terribly parity calibrated, which corroborates our argument from Section~\ref{sec:counter-examples}, that calibration in regression does not imply parity calibration.
In the same plot, we can see that MW, IW and OPS are all able to improve parity calibration, but the numerical results in Table~\ref{tab:pressure_numerical} show that OPS produces superior parity probabilities w.r.t. all of the metrics considered.

\begin{figure}[ht]
\begin{center}
\centerline{\includegraphics[width=\columnwidth]{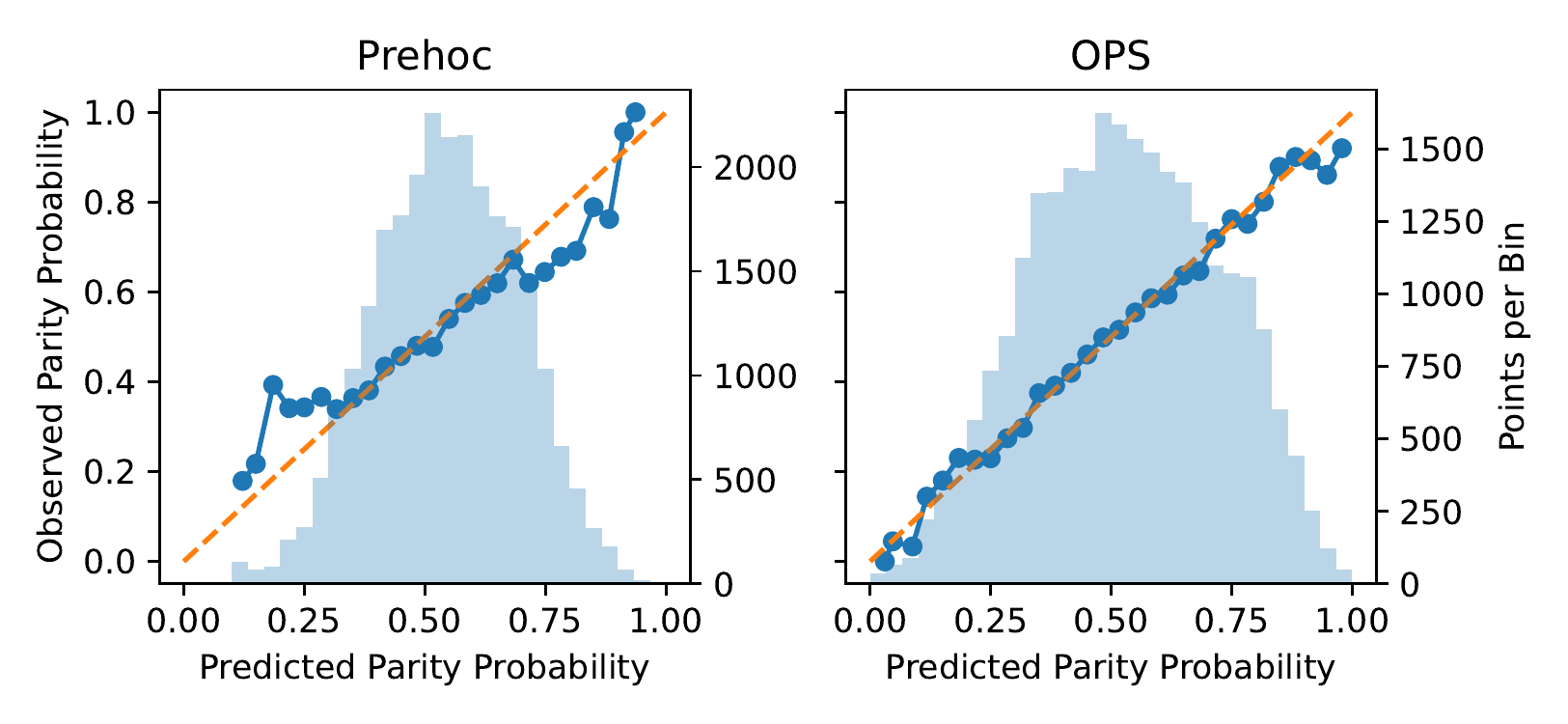}}
\caption{
The base classification model for Pressure (Prehoc) is better parity calibrated than the base regression model (Figure~\ref{fig:pressure-gaussian-base} Prehoc), but OPS still improves its parity calibration and sharpness.
}
\label{fig:pressure-binary-prehoc-ops}
\end{center}
\vspace{-8mm}
\end{figure}

\begin{figure*}[t]
\centering 
\includegraphics[width=0.93\textwidth]{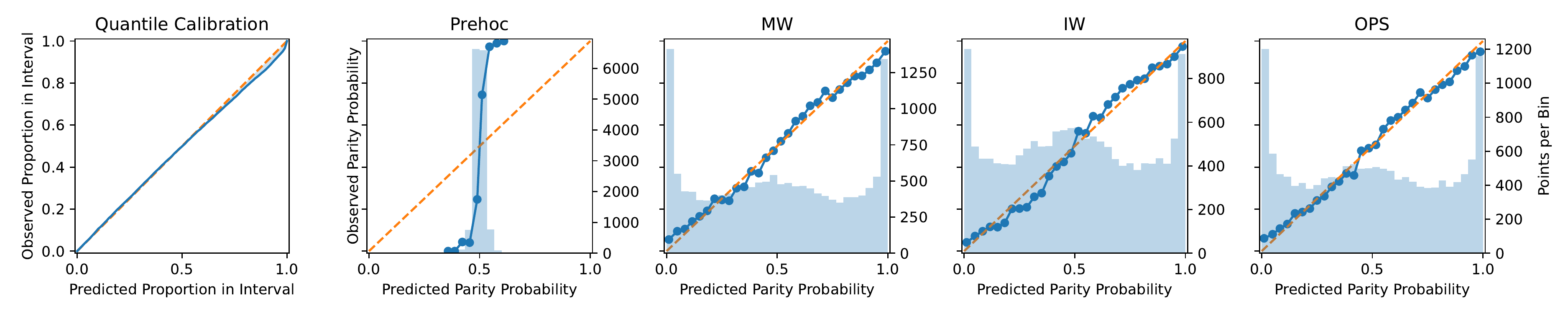}
\caption{All methods (MW, IW, OPS) perform equally well in calibrating the Prehoc parity probabilities of the nuclear fusion dynamics model. The left two plots display the quantile calibration and parity calibration of the base dynamics model.}
\label{fig:fusion_parity}
\end{figure*}

\begin{table*}[t]
\centering
{\setlength{\tabcolsep}{5pt}
\begin{tabular}{lccccc}
\toprule
          & QCE $\downarrow$ & PCE $\downarrow$ & Sharp $\uparrow$  & Acc $\uparrow$ & AUROC $\uparrow$\\ 
\midrule
Prehoc  & \textbf{0.0108$\pm$0.0003} & 0.2571$\pm$0.0003 & 0.3243$\pm$0.0002 & \textbf{0.7727$\pm$0.0003} & \textbf{0.8536$\pm$0.0002} \\
MW      & N/A                        & 0.0266$\pm$0.0002 & 0.3345$\pm$0.0002 & 0.7665$\pm$0.0003 & 0.8463$\pm$0.0002 \\
IW      & N/A                        & 0.0291$\pm$0.0002 & \textbf{0.3385$\pm$0.0002} & \textbf{0.7726$\pm$0.0003} & \textbf{0.8533$\pm$0.0002} \\
OPS     & N/A                        & \textbf{0.0261$\pm$0.0002} & 0.3334$\pm$0.0002 & 0.7629$\pm$0.0002 & 0.8440$\pm$0.0002 \\
\bottomrule
\end{tabular}}
\caption{
MW, IW, and OPS all improve parity calibration and sharpness of the Prehoc fusion dynamics model predicting $\beta_N$, 
while maintaining roughly the same level of accuracy. $\pm$ indicates mean $\pm$ 1 standard error, across 50 test trials.
Best value for each metric is in bold.
}
\label{tab:fusion_numerical}
\vspace{-3mm}
\end{table*}

\textbf{Binary classifiers as expert forecasters.}
While we have so far assumed that the expert forecaster provides regression models $\hF_t$, one may argue that an expert forecaster may be well-aware that the downstream user is primarily concerned with parity probabilities. Accordingly, the expert may choose to directly model parity probabilities in the context of a binary classification problem.

In Figure~\ref{fig:pressure-binary-prehoc-ops} and Table~\ref{tab:pressure-binary-prehoc-ops}, we show results from training a base binary classifier with 
parity outcome labels and applying posthoc calibration.
As expected, the prehoc parity probabilities of the binary classification model is significantly better calibrated than the regression model.
Posthoc calibration still improves parity calibration further, especially in the case of OPS. In fact, OPS is the only method which improves all of the metrics simultaneously, while MW and IW notably worsen sharpness and AUROC. The full set of reliability diagrams is provided in Figure~\ref{fig:pressure-binary-base-full-comparison} in Appendix~\ref{app:weather-additional-results}.

\textbf{Results across all 7 timeseries.}
Table~\ref{tab:weather-average} in Appendix~\ref{app:weather-additional-results} shows each metric averaged across all 7 prediction targets: Table~\ref{tab:weather-average-gaussian-base} displaying results with the base regression model, and \ref{tab:weather-average-binary-base} that of the base classification model. The pattern observed for the Pressure timeseries tend to hold on average across all 7 timeseries.

\begin{figure}[t]
\begin{center}
\centerline{\includegraphics[width=\columnwidth]{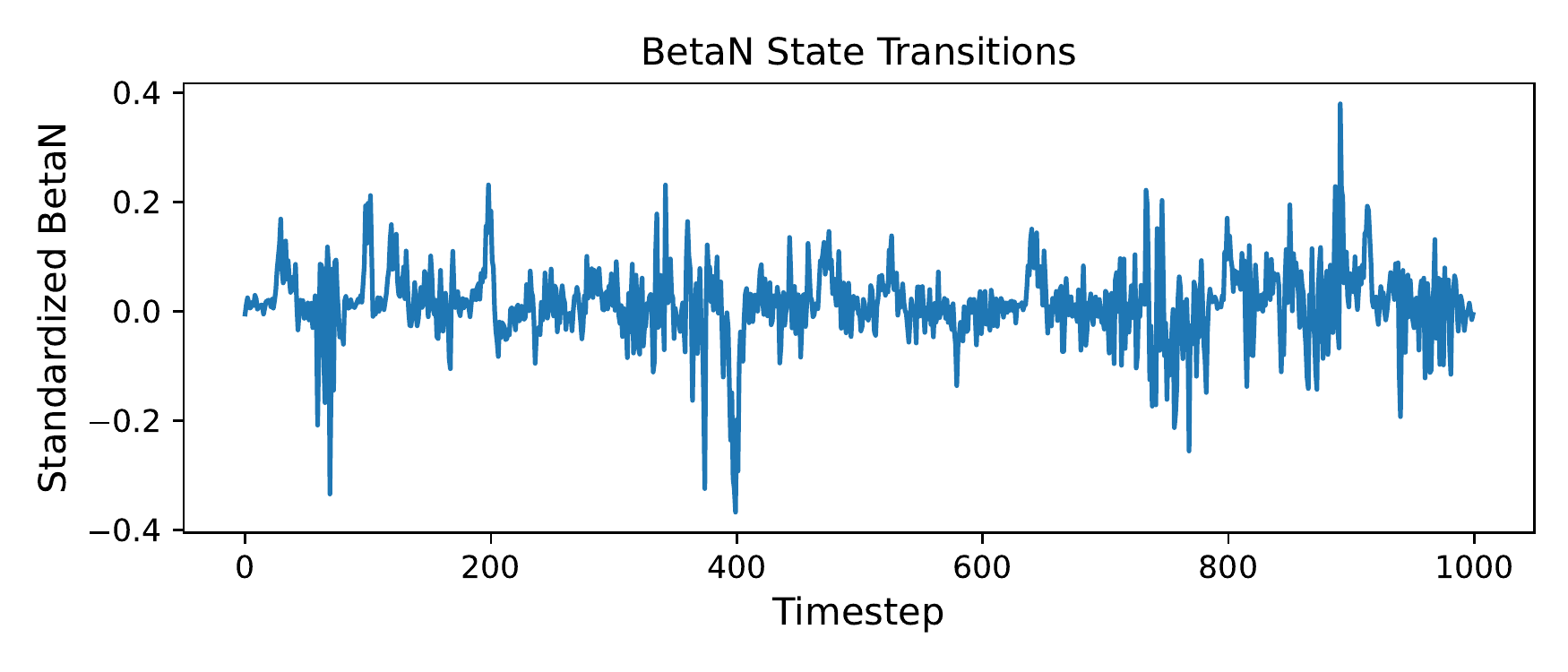}}
\caption{State transitions of the $\beta_N$ signal during nuclear fusion experiments (``shots'') concatenated across 50 training shots resemble trend-less noise.}
\label{figure:betan_timeseries}
\end{center}
\vspace{-9mm}
\end{figure}

\subsection{Case Study 3: Model-based Control for nuclear fusion}
\label{sec:fusion-case-study}
Nuclear fusion is the physical process during which atomic nuclei combine together to form heavier atomic nuclei, while releasing atomic particles and energy.
Although fusion is possibly a safe, clean, and fuel-abundant technology for the future~\citep{morse2018nuclear}, there are various challenges to realizing fusion power, one of which is controlling nuclear fusion reactions~\citep{humphreys2015novel}.

Recently, model-based control methods,
where a dynamics model of the system is learned and used 
to optimize control policies, has emerged as an effective control method for fusion devices~\citep{abbate2023general}.
To the experimenter utilizing the dynamics model, 
it is of significant interest to know when certain signals 
will increase, and whether the dynamics model 
assigns correct probabilities to the events~\citep{char2021model}.
In this section, we consider the problem of 
predicting the parity of $\beta_N$, which is a signal indicating reaction efficiency in a fusion device called a tokamak.

To this end, we design our empirical case study as follows.
We take a pretrained dynamics model which was trained with a logged database of $10294$ fusion experiments (referred to as ``shots'') conducted on the DIII-D tokamak~\citep{luxon2002design}, a device in San Diego, CA, USA.
This pretrained model has been used for model-based policy optimization for deployment in actual fusion experiments on this device~\citep{char2021model, seo2021feedforward, abbate2021data}.
The model architecture is a recurrent probabilistic neural network (RPNN), which is a recurrent neural network with a Gaussian output head. We refer the reader to 
Appendix~\ref{app:fusion-experiment-details} for more details of the dynamics model and dataset.
For testing, we allocate a set of 900 held-out test shots.
On this test set, we produce the model's distributional predictions for $\beta_N$ as the expert forecast.
We concatenate the forecasts and the actual observed $\beta_N$ values across the $900$ test shots in chronological order into a single timeseries to assess parity calibration.

Figure~\ref{fig:fusion_parity} and Table~\ref{tab:fusion_numerical} indicate that the expert forecast (Prehoc) is quantile calibrated
but parity miscalibrated. The accuracy metrics in Table~\ref{tab:fusion_numerical} indicate that despite prehoc's poor parity calibration, the model is still highly predictive, with an AUROC $>0.85$.
MW, IW and OPS significantly improve parity calibration and sharpness, while maintaining roughly the same level of accuracy.

We note that the $\beta_N$ timeseries, as displayed in Figure~\ref{figure:betan_timeseries}, tends to fluctuate rapidly, between timesteps and between shots, almost resembling white noise.
The pretrained model still manages to model the signal well,
and assigns correct tendencies of increases/decreases in $\beta_N$: the relibility diagram of prehoc in Figure~\ref{fig:fusion_parity} shows that although the parity probabilities are not aligned with the empirical frequencies,
they predict higher probabilities for actually higher frequency events. 
We believe this provides for a relatively easy posthoc calibration problem, thus all methods (MW, IW, OPS) perform equally well.
Hence, this case study highlights the significance of the base model's initial parity probabilities, especially in alleviating the difficulty of posthoc calibration.

\section{Conclusion}
We considered the problem of forecasting whether a continuous-valued sequence is going to increase or decrease at the next time step. Such scenarios, where relative changes are more interpretable than actual values, are ubiquitous: COVID-19 cases per day, weather, or stock prices. 
In this context, we proposed the notion of parity calibration. To be parity calibrated, a forecaster must predict probabilities for the outcome increasing at the next time step, and these probabilities should be calibrated in the binary sense. 

A decision-maker may attempt to achieve parity calibration by using regression forecasts produced by an expert forecaster. However, this is unlikely to give parity calibration. Instead, we proposed the usage of posthoc binary calibration techniques to achieve parity calibration. 
Specifically, we advocated for a recently proposed online Platt scaling algorithm (OPS) in this setting. 
In three real-world empirical case studies, OPS consistently improves the overall quality of parity probabilities compared to the expert forecaster.

\begin{contributions}
    YC led the project as first author.
    AR played a key role in acquiring and interpreting the COVID-19 data for our experiments and contributed to discussions during project development.
    CG played the advisory role. He contributed the initial idea, planned the project direction, and steered the execution of the initial write-up as well as consequent revisions.
\end{contributions}

\begin{acknowledgements}
    We would like to thank our Ph.D. advisors---Jeff Schneider (YC), Roni Rosenfeld (AR), and Aaditya Ramdas (CG)---for enabling us to pursue this student-only work. 
    YC is supported in part by US Department of Energy grants under contract numbers DE-SC0021414 and DE-AC02-09CH1146, and the Kwanjeong Educational Foundation. AR is supported by McCune Foundation grant FP00004784. CG is supported by the Bloomberg Data Science Ph.D. Fellowship. 
    The authors would also like to thank the anonymous UAI reviewers for their valuable feedback.
\end{acknowledgements}

\bibliography{chung_631}

\title{Parity Calibration\\(Supplementary Material)}
  
\onecolumn 
\hypersetup{urlcolor=black}
\maketitle
\renewpagestyle{plain}{%
    \setfoot{}{\thepage}{}
}
\hypersetup{urlcolor=NavyBlue}

\appendix
\section{Details on Evaluation: reliability diagrams and metrics}
We provide details on how we assess a sequence of distributional forecasts $\{\hF_t\}_{t=1}^{T}$ and parity probabilities $\{\hp_t\}_{t=1}^{T}$, given a
test dataset $\mathcal{D}_{\text{test}} = \{\x_t, y_t\}_{t=1}^{T}$. We assess distributional forecasts via Quantile Calibration, and the parity probabilities via Parity Calibration, Sharpness, and Accuracy metrics.
\label{app:metrics}
\begin{itemize}
    \item \textbf{Quantile Calibration: reliability diagram and calibration error}
    
    To assess the quantile calibration of the distributional forecast $\hF_t$, we produce the reliability diagram using the \textit{Uncertainty Toolbox}~\citep{chung2021uncertainty}. This process works as follows. We take $100$ equi-spaced quantile levels in $[0, 1]$: $p_i \in$ \texttt{np.linspace(0, 1, 100)},
    and for each $p_i$, we compute the empirical coverage of the predictive quantile $\hF^{-1}_t(p_i)$ 
    with $\frac{1}{T}\sum_{t=1}^{T}\indicator{y_t \leq \hF^{-1}_t(p_i)}$, and we denote this quantity as $p_{i, \text{obs}}$.
    Note that $p_{i, \text{obs}}$ is an empirical estimate of the term $\frac{1}{T}\sum_{t=1}^{T}F_t(\hF^{-1}_t(p_i))$, from Eq.~\eqref{eq:probabilistic-calibration}.
    The reliability diagram is produced by plotting $\{p_i\}$ on the $x$-axis against $\{p_{i, \text{obs}}\}$ on the $y$-axis. Quantile Calibration Error (QCE) is then computed as the average of the absolute difference between $p_i$ and $p_{i, \text{obs}}$ over the $100$ values of $p_i$: $\frac{1}{100}\sum_{i=1}^{100}\mid p_{i, \text{obs}} - p_i\mid$.
    
    \item \textbf{Parity Calibration: reliability diagram and calibration error} 

    For parity calibration, we produce the reliability diagram following the standard method in binary classification~\citep{degroot1981assessing, niculescu2005predicting}. Note that the parity probability $\hp_t$ is a prediction for the parity outcome $\widetilde{y}_t := \indicator{y_t \leq y_{t-1}}$ (Eq.~\eqref{eq:parity-outcome}). Specifically, we first take $30$ fixed-width bins of the predicted parity probabilities: $\{B_{m}\}_{m=1}^{30}$, where $B_{m} = [\frac{m-1}{30}, \frac{m}{30})$ for $m < 30$ and $B_{30} = [\frac{29}{30}, 1]$.
    The average outcome in bin $B_m$ is computed as $\text{obs}(B_m) = \frac{1}{\mid B_m \mid} \sum_{t: \hp_t \in B_m} \indicator{\widetilde{y}_t = 1}$, and the average prediction of bin $B_m$ is computed as $\text{pred}(B_m) = \frac{1}{\mid B_m \mid} \sum_{t: \hp_t \in B_m} \hp_t$.
    The reliability diagram is then produced by plotting $\text{pred}(B_m)$ on the $x$-axis against $\text{obs}(B_m)$ on the $y$-axis.
    The blue bars in the background of each parity calibration reliability diagram represents the size of the bin: $|B_m |$. Parity Calibration Error (PCE) is then computed with this reliability diagram following the standard definition of ($\ell_1$-)expected calibration error (ECE): $\sum_{m=1}^{30}\frac{\mid B_m \mid}{T}\mid \text{obs}(B_m) - \text{pred}(B_m) \mid$.
    
    \item \textbf{Sharpness}
    
    Assuming the same notation as above, sharpness is computed as: $\sum_{m=1}^{M}\frac{\mid B_m \mid}{T}\cdot \text{obs}(B_m)^{2}$, where $M$ is the total number of bins. As indicated above, we use $M=30$ in all of our experiments. 
    We provide some additional intuition on this metric. A perfectly knowledgeable forecaster which outputs $\hat{p}_t = \widetilde{y}_t$ will place all predictions in either $B_1$ or $B_M$ and achieve sharpness $ = \frac{\mid B_1 \mid}{T}\cdot \text{obs}(B_1)^{2} + \frac{\mid B_M \mid}{T}\cdot \text{obs}(B_M)^{2} = \frac{\mid B_1 \mid}{T}\cdot 0^{2} + \frac{\mid B_M \mid}{T}\cdot 1^{2} = \frac{\mid B_M \mid}{T} = \frac{\sum_{t=1}^{T} \widetilde{y}_t}{T}$. On the other hand, if the forecaster places all predictions into a single bin $B_k$, then its sharpness will be $\text{obs}(B_k)^{2} = \left(\frac{\sum_{t=1}^{T} \widetilde{y}_t}{T}\right)^{2}$.
    It can be shown that sharpness is always within the closed interval $\left[ \left(\frac{\sum_{t=1}^{T} \widetilde{y}_t}{T}\right)^{2}, \frac{\sum_{t=1}^{T} \widetilde{y}_t}{T} \right]$~\citep{brocker2009reliability}. 
    Intuitively, sharpness measures the degree to which the forecaster attributes different valued predictions to events with different outcomes (i.e. labels). Hence, a sharper, or more precise, forecaster has more discriminative power, and this is reflected in a higher sharpness metric.
     
    \item \textbf{Accuracy metrics (Acc and AUROC)} 
    
    Accuracy is measured in the binary classification sense, where the true labels are the observed parity outcomes: $\indicator{y_t \leq y_{t-1}}$ (Eq.~\eqref{eq:parity-outcome}).
    \begin{itemize}
        \item \textbf{Binary accuracy (Acc)} is computed by regarding $\hp_t \geq 0.5$ as the positive class prediction, and the opposite case as the negative class prediction. 
        \item \textbf{Area under the ROC curve (AUROC)} is computed using the \texttt{scikit-learn} Python package, which implements the standard definition of the score. 
        Specifically, we called the function \texttt{sklearn.metrics.roc\_auc\_score} with the predictions $\{\hp_t\}$ and labels $\indicator{y_t \leq y_{t-1}}$.
    \end{itemize}
\end{itemize}

\section{Additional Details on Case Studies}
\subsection{Additional Details on COVID-19 Case Study}\label{app:covid-appendix}
\subsubsection{Details on Interpolating Expert Forecasts for COVID-19 Case Study}
\begin{figure}[h]
\begin{center}
\includegraphics[width=0.4\textwidth]{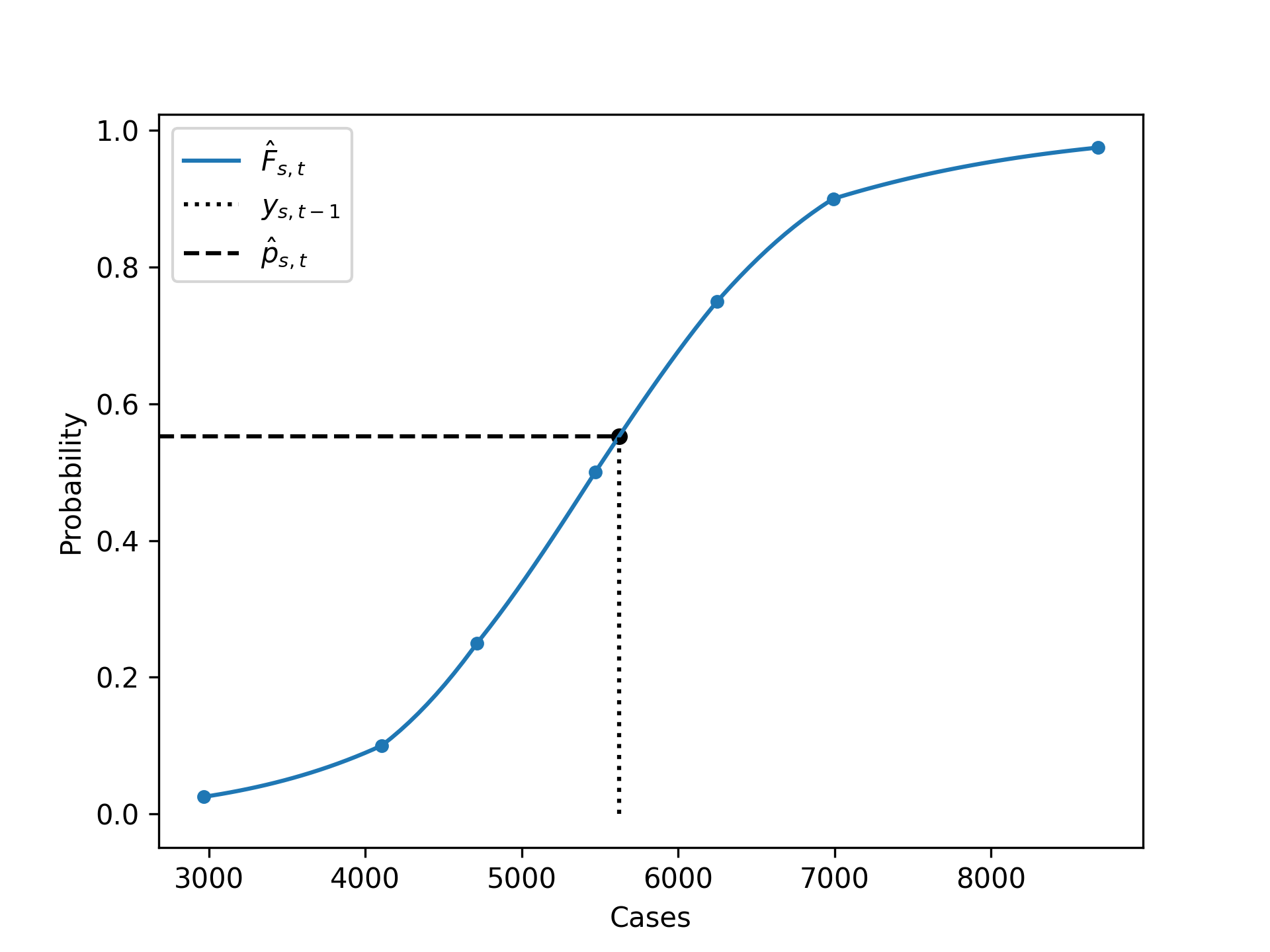}
\caption{We use a piece-wise Gaussian interpolation of the expert forecast quantiles to estimate the predictive cdf, from which we then calculate the parity probabilities.}
\label{fig:covid-interp-parity-prob}
\end{center}
\vskip -0.2in
\end{figure}

The expert forecast provided by the COVID-19 Forecast Hub is represented as a set of quantiles.
To derive the parity probabilities $\hp_{s,t}$, we need to interpolate the expert forecast, as the forecast contains predicted quantiles at only 7 quantile levels : $\{0.025, 0.1, 0.25, 0.5, 0.75, 0.9, 0.975\}$. 
We interpolate under the assumption that the density between two adjacent quantiles $\tau_k$ and $\tau_{k+1}$ are defined by the normal distribution specified by those two quantiles. Specifically, for two quantiles $\tau_k$ and $\tau_{k+1}$ and forecast values $x^{(s,t)}_k$ and $x^{(s,t)}_{k+1}$, we compute $$\sigma^{(s,t)}_k = \frac{x^{(s,t)}_{k+1} - x^{(s,t)}_k}{\Phi^{-1}(\tau_{k+1}) - \Phi^{-1}(\tau_k)},$$
$$\mu^{(s,t)}_k = x^{(s,t)}_k - \sigma^{(s,t)}_k \Phi^{-1}(\tau_k),$$
where $\Phi$ is the standard normal cdf. For each forecast, if $x^{(s,t)}_k \leq y_{s,t-1} < x^{(s,t)}_{k+1}$, then the parity probability $$\hp_{s,t}= \Phi\left(\frac{y_{s,t-1} - \mu^{(s,t)}_k}{\sigma^{(s,t)}_k}\right).$$
If $y_{s,t-1} < x^{(s,t)}_1$, we can extrapolate using $\mu^{(s,t)}_1$ and $\sigma^{(s,t)}_1$, and if $y_{s,t-1} >= x^{(s,t)}_7$, we can extrapolate using $\mu^{(s,t)}_6$ and $\sigma^{(s,t)}_6$. However, this never occurs with the forecasts and observations in this dataset.
Figure~\ref{fig:covid-interp-parity-prob} provides a visualization of this interpolation scheme.

\subsubsection{Details on Experiment Setup for COVID-19 Case Study}
Section~\ref{sec:covid-single-timeseries} compares the expert forecaster, its parity probabilities and posthoc calibration by OPS. 
We did not tune OPS hyperparameters in this experiment, so the full 119 weeks' worth of data was used for testing and reporting the results.

For Section~\ref{sec:covid-methods-comparison}, the first 20 weeks' worth of data was used for tuning hyperparameters, and the reported results are based on the remaining 99 weeks' worth of data as the test set.

For the decision-making experiment in Section~\ref{sec:covid-decision-making}, we used the parity probabilities produced from Section~\ref{sec:covid-methods-comparison}.\\
Although the chosen loss function is just one example, we observe that similar results hold with any loss function that satisfies: $l_{2, 3}\leq l_{2, 2} \leq l_{1, 1} \leq l_{2, 1} \leq l_{1, 2} \leq l_{1, 3}$.

\subsection{Additional Details on Weather Forecasting Case Study}\label{app:weather-appendix}
\subsubsection{Details on Experiment Setup for Weather Forecasting Case Study}\label{app:weather-experiment-details}
We used the modeling and training infrastructure 
provided by the Keras tutorial on \textit{Timeseries Forecasting for Weather Prediction}\footnote{\url{https://keras.io/examples/timeseries/timeseries_weather_forecasting/}} which models this same dataset with an LSTM network~\citep{hochreiter1997long}. 
We made one change to the model provided by the tutorial: since we are interested in probabilistic forecasts instead of point forecasts, 
we changed the head of the model and the loss function from a point output trained with mean squared error loss 
to a mean and variance output that parameterizes a Gaussian distribution 
and trained it with the Gaussian likelihood loss.
Such a model is also referred to as a mean-variance network or a 
probabilistic neural network  \citep{lakshminarayanan2017simple, nix1994estimating}, 
and it is one of the most popular methods currently used in probabilistic regression.

While the tutorial's setup takes as input the past 120 hours' window of 7 features to predict the value of one feature (Temperature) 12 hours into the future, 
we expand the setting to predict all 7 features: Pressure, Temperature, Saturation vapor pressure, Vapor pressure deficit, Specific humidity, Airtight, and Wind speed.
We thus train 7 separate base regression models, one for each prediction target.

For the in-text experiment \textbf{Binary classifers as expert forecasts}, we trained binary classification base models with parity outcomes (Eq.~\eqref{eq:parity-outcome}) as the labels and took this model as the expert forecaster. 
We adopted the same model architecture as the base regression model and changed the last layer to output a logit. We then trained the model with the cross entropy loss.

The full Jena dataset spans from the beginning of January 2009 to the end of December 2016, with $420,551$ datapoints in total. In chronological order, we set $272,638$ datapoints to train the base models (both the regression and classification model) and the subsequent $83,390$ datapoints for validation. Following the same model training procedure as the tutorial, training was stopped early if the validation loss did not increase for 20 training epochs.

Afterwards, in running the posthoc calibration methods (MW, IW, and OPS), we used the last $8,640$ datapoints of the validation set to tune the hyperparameters of each calibration method, and used subsequent windows of $8,640 \times 3 = 25,920$ datapoints for testing.

We run 50 test trials with a moving test timeframe to produce the mean and standard errors reported in Tables~\ref{tab:pressure_numerical} and \ref{tab:pressure-binary-prehoc-ops}. Denoting the first test window as $[t+1, t+H]$ (i.e. $H$ is set to $25,920$), we move this frame by a multiple of a fixed offset $c$ into the future, and repeat this 50 times, to create a new set of 50 test sets. The resulting new test timeframes are $[t+1+(ck), t+H+(ck)]$, where $k = 0, 1, 2, \dots 49$, and $c$ was set to $336$.

\subsubsection{Additional Results on Weather Forecasting Case Study}\label{app:weather-additional-results}

We shows additional plots and tables from the experimental results in Section~\ref{sec:weather-case-study} of the main paper.

\begin{figure}[ht]
    \centering
    \includegraphics[width=\linewidth]{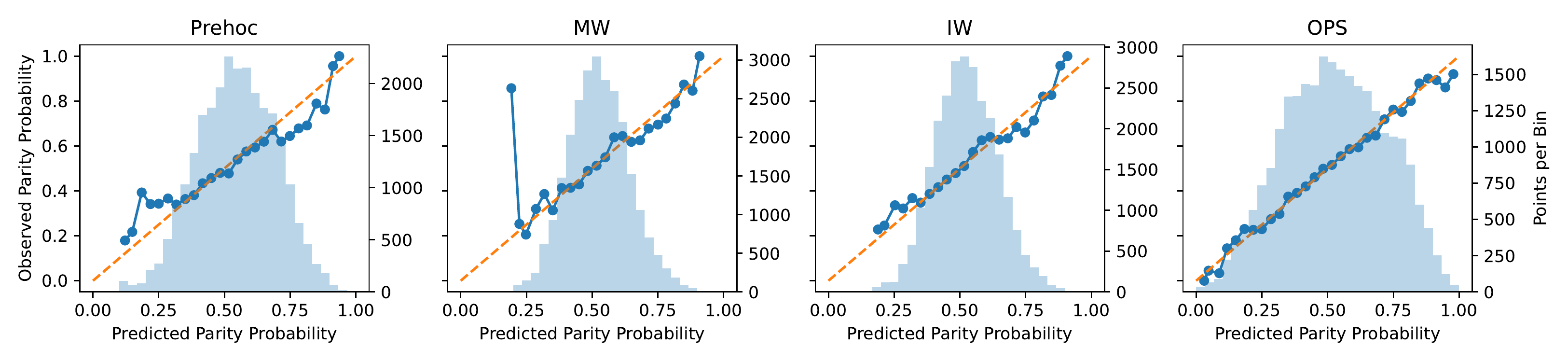}
    \caption{Reliability diagrams with a binary classification base model predicting Pressure. This is the full set of reliability diagrams for Figure~\ref{fig:pressure-binary-prehoc-ops} from Section~\ref{sec:weather-case-study}. The left-most plot shows parity calibration of the base classification model (Prehoc), and the next three plots show the effects of MW, IW and OPS in calibrating the Prehoc parity probabilities. OPS produces the most calibrated and sharp parity probabilities.}
    \label{fig:pressure-binary-base-full-comparison}
\end{figure}

Figure~\ref{fig:pressure-binary-base-full-comparison} displays the full set of reliability diagrams for Figure~\ref{fig:pressure-binary-prehoc-ops}, which corresponds to the in-text experiment \textbf{Binary classifiers as expert forecasts} in Section~\ref{sec:weather-case-study}.

Table~\ref{tab:weather-average} displays the numerical results from the weather forecasting case study when averaged across all 7 prediction target settings. This corresponds to the in-text experiment \textbf{Results across all 7 timeseries} in Section~\ref{sec:weather-case-study}. To produce these results, we fixed the test timeframe to be the first test timeframe $[t+1, t+H]$ for all prediction target settings, then computed the mean and standard errors across the 7 sets of metrics produced (one set for each prediction target).

\begin{table*}[ht!]
\centering
\vspace{2mm}
\begin{subcaptionblock}{\textwidth}
    \centering
    \begin{tabular}{lccccc}
    \toprule
        & QCE $\downarrow$ & PCE $\downarrow$ & Sharp $\uparrow$ & Acc $\uparrow$ & AUROC $\uparrow$ \\ \midrule
    Prehoc  & $\mathbf{0.0266 \pm 0.0052}$ & $0.2794 \pm 0.0161$ & $0.2915 \pm 0.0117$ & $0.4902 \pm 0.0159$ & $0.4806 \pm 0.0249$\\ 
    MW  & N/A & $0.0233 \pm 0.0048$ & $0.2913 \pm 0.0117$ & $0.5610 \pm 0.0106$ & $0.5419 \pm 0.0195$\\
    IW  & N/A & $0.0188 \pm 0.0047$ & $0.2913 \pm 0.0118$ & $0.5630 \pm 0.0099$ & $0.5403 \pm 0.0209$\\ 
    OPS  & N/A & $\mathbf{0.0159 \pm 0.0009}$ & $\mathbf{0.2961 \pm 0.0122}$ & $\mathbf{0.5790 \pm 0.0122}$ & $\mathbf{0.5830 \pm 0.0217}$\\
    \bottomrule
    \end{tabular}
    \caption{Numerical results averaged across all 7 prediction settings where the base model is a Gaussian regression model. The base regression model (Prehoc) tends to be well quantile calibrated (QCE) but terribly parity calibrated (PCE). All methods (MW, IW, OPS) improve parity calibration, but OPS is the only method which improves all metrics simultaneously. Best value for each metric is in bold.}
    \label{tab:weather-average-gaussian-base}
\end{subcaptionblock}

\vspace{3mm}
\begin{subcaptionblock}{\textwidth}
    \centering
    \begin{tabular}{lcccc}
    \toprule
             & PCE $\downarrow$ & Sharp $\uparrow$ & Acc $\uparrow$ & AUROC $\uparrow$\\ \midrule
    Prehoc  & $0.0247 \pm 0.0016$ & $0.3049 \pm 0.0074$ & $0.6078 \pm 0.0099$ & $0.6348 \pm 0.0136$\\
    MW  & $0.0170 \pm 0.0018$ & $0.3049 \pm 0.0075$ & $0.6061 \pm 0.0102$ & $0.6340 \pm 0.0143$\\ 
    IW  & $0.0156 \pm 0.0012$ & $0.3047 \pm 0.0074$ & $0.6075 \pm 0.0098$ & $0.6340 \pm 0.0136$\\ 
    OPS  & $\mathbf{0.0135 \pm 0.0013}$ & $\mathbf{0.3134 \pm 0.0075}$ & $\mathbf{0.6278 \pm 0.0121}$ & $\mathbf{0.6643 \pm 0.0183}$\\ 
    \bottomrule
    \end{tabular}
    \caption{Numerical results averaged across all 7 prediction settings where the base model is a binary classification model trained with parity outcome labels. The base classification model (Prehoc) tends to be much better parity calibrated than when a regression base model is used (above Table~\ref{tab:weather-average-gaussian-base}). All methods (MW, IW, OPS) improve parity calibration further, but OPS is the only method which improves all metrics simultaneously. Notably, MW and IW tends to decrease the accuracy of the parity probabilities. Best value for each metric is in bold.}
    \label{tab:weather-average-binary-base}
\end{subcaptionblock}
\caption{Numerical results from the weather forecasting case study (Section~\ref{sec:weather-case-study}), averaged across all 7 forecasting targets. Table~\ref{tab:weather-average-gaussian-base} displays results with the Gaussian regression base model, and Table~\ref{tab:weather-average-binary-base} displays results with the binary classification base model. $\pm$ indicates mean $\pm$ 1 standard error, across the 7 prediction target settings.}
\label{tab:weather-average}
\vspace{-4mm}
\end{table*}

\subsection{Additional Details on Control in Nuclear Fusion Case Study}\label{app:fusion-appendix}
\subsubsection{Details on Experiment Setup for Control in Nuclear Fusion Case Study}\label{app:fusion-experiment-details}
The expert forecaster for the nuclear fusion experiment in Section~\ref{sec:fusion-case-study} is provided by a pretrained dynamics models that was used to optimize control policies for deployment 
on the DIII-D tokamak~\citep{luxon2002design}, a nuclear fusion device in San Diego that is operated by General Atomics.
The dynamics model was trained with logged data from past experiments (referred to as ``shots'') on this device. Each shot consists of a trajectory of (state, action, next state) transitions, and one trajectory consists of $\sim20$ transitions (i.e. $20$ timesteps).

As input, the model takes the current state of the plasma and the actuator settings (i.e. actions). The model outputs a multi-dimensional predictive distribution over the state variables in the next timestep. The state is represented by three signals: $\beta_N$ (the ratio of plasma pressure over magnetic pressure), \textit{density} (the line-averaged electron density), and \textit{li} (internal inductance).
For the actuators, the model takes in the amount of power and torque injected from the neutral beams, the current, the magnetic field, and four shape variables (\textit{elongation}, $a_{minor}$, \textit{triangularity-top}, and \textit{triangularity-bottom}). This, along with the states, makes for an input dimension of 11 and output dimension of 3 for the states.

The model was implemented with a recurrent probabilistic neural network (RPNN), which features an encoding layer by an RNN with 64 hidden units followed by a fully connected layer with 256 units, and a decoding layer of fully connected layers with [128, 512, 128] units, which finally outputs a 3-dimensional isotropic Gaussian parameterized by the mean and a log-variance prediction.

The training dataset consisted of trajectories from 10294 shots, and the model was trained with the Gaussian likelihood loss, with a learning rate of 0.0003 and weight decay of 0.0001.
In using dynamics models to sample trajectories and train policies, the key metric 
practitioners are concerned with is explained variance, hence explained variance on a held out validation set of 1000 shots was monitored during training. Training was stopped early if there was no improvement in explained variance over the validation set for more than 250 epochs.
The test dataset consisted of another held-out set of 900 shots, with which we report all results presented in Section~\ref{sec:fusion-case-study}.

In all of our experiments, since $\beta_N$ is the key signal of interest in our problem setting, we just examine the predictive distribution for $\beta_N$ in the model outputs and ignore the other dimensions of the outputs. 

In running the posthoc calibration methods (MW, IW, and OPS), we used the same validation set to tune the hyperparameters of each calibration method, and used windows of $15,000$ datapoints from the concatenated test shot data for testing.

We run 50 test trials with a moving test timeframe to produce the mean and standard errors reported in Table~\ref{tab:fusion_numerical}. Denoting the first test window as $[t+1, t+H]$ (i.e. $H$ is set to $15,000$), we move this frame by a multiple of a fixed offset $c$ into the future, and repeat this 50 times, to create a set of 50 test datasets. The resulting test timeframes are $[t+1+(ck), t+H+(ck)]$, where $k = 0, 1, 2, \dots 49$, and $c$ was set to $100$.

\section{Details on Hyperparameters}
\label{app:hyperparameters}

Each of the three calibration methods we consider in Section~\ref{sec:main-parity-calibration-methodology}, which we use in our experiments in Section~\ref{sec:experiments}, requires a set of hyperparameters.
\begin{itemize}
    \item \textbf{MW} requires \texttt{uf} and \texttt{ws}.
    \begin{itemize}
        \item \texttt{uf} determines how often the PS parameters $(a^{\text{MW}}, b^{\text{MW}})$ are updated.
        \item \texttt{ws} determines the size of the calibration set that is used to update the PS parameters
    \end{itemize}
    \item \textbf{IW} requires \texttt{uf}.
    \begin{itemize}
        \item \texttt{uf} determines how often the PS parameters $(a^{\text{IW}}, b^{\text{IW}})$ are updated.\\
        Note that  IW always uses all of the data seen so far to update the PS parameters.
    \end{itemize}
    \item \textbf{OPS} requires \texttt{$\gamma$} and \texttt{D}.
    \begin{itemize}
        \item \texttt{$\gamma$} can be understood as step size for the OPS updates.
        \item \texttt{$D$} can be understood as regularization for the OPS updates.
    \end{itemize}
\end{itemize}
We provide details on how these hyperparameters were tuned for each of the three case studies.
\subsection{Hyperparameters for COVID-19 Case Study}
We observed that OPS performed well with the default hyperparameters, 
so we did not tune hyperparameters for OPS for the COVID-19 case study.
The default hyperparameter values used for OPS were $\gamma = 0.001$ and $\texttt{D} = 10$.

For MW and IW,
we tuned hyperparameters by optimizing parity calibration error (PCE, Section~\ref{sec:experiments}) on the first 20 weeks' worth of data as the validation set, over the following grids:
\begin{itemize}
    \item \texttt{uf} $\in [1, 2, 3, 4, 5, 6, 7, 8, 9, 10]$, separately for MW and IW
    \item \texttt{ws} $\in [1, 2, 3, 4, 5, 6, 7, 8, 9, 10]$, for MW.
\end{itemize}
The COVID-19 dataset records data for each week, so the grid size of 1 represents 1 week.

The tuned hyperparameters we used for MW and IW are as follows:
\begin{itemize}
    \item MW: $\texttt{uf}=1, \texttt{ws}=10$
    \item IW: $\texttt{uf}=5$
\end{itemize}

\subsection{Hyperparameters for Weather Forecasting Case Study}
For each calibration method, the hyperparameters were tuned by optimizing parity calibration error (PCE, Section~\ref{sec:experiments}) on the validation dataset over the following grids:
\begin{itemize}
    \item \texttt{uf} $\in [1, 24, 168, 336, 720, 2160]$, separately for MW and IW
    \item \texttt{ws} $\in [24, 168, 336, 720, 2160, 4320, 8640]$, for MW
    \item $\gamma$ $\in$ [1e-5, 5e-5, 1e-4, 5e-4, 1e-3, 5e-3, 1e-2], for OPS
    \item \texttt{D} $\in [1, 10, 30, 50, 70, 100, 150, 200]$, for OPS.
\end{itemize}
The hyperparameters were tuned separately for each base model setting (regression and classification), for each method (MW, IW, and OPS), and for each base model predicting one of 7 targets (Pressure, Temperature, Saturation vapor pressure, Vapor pressure deficit, Specific humidity, Airtight, and Wind speed).

The tuned hyperparameters we used are as follows:
\begin{itemize}
    \item \textbf{Base Regression Model}
    \begin{itemize}
        \item Pressure Model
        \begin{itemize}
            \item MW: $\texttt{uf}=2160 , \texttt{ws}=8640 $
            \item IW: $\texttt{uf}=2160$
            \item OPS: $\gamma=1\text{e-5}, \texttt{D}=50$
        \end{itemize}
        \item Temperature Model
        \begin{itemize}
            \item MW: $\texttt{uf}=336 , \texttt{ws}=8640 $
            \item IW: $\texttt{uf}=168$
            \item OPS: $\gamma=1\text{e-5} , \texttt{D}=30 $
        \end{itemize}
        \item Saturation Vapor Pressure Model
        \begin{itemize}
            \item MW: $\texttt{uf}=2160 , \texttt{ws}=2160 $
            \item IW: $\texttt{uf}=336$
            \item OPS: $\gamma=1\text{e-4} , \texttt{D}=10 $
        \end{itemize}
        \item Vapor Pressure Deficit Model
        \begin{itemize}
            \item MW: $\texttt{uf}=1 , \texttt{ws}=4320 $
            \item IW: $\texttt{uf}=1$
            \item OPS: $\gamma=1\text{e-3} , \texttt{D}=1$
        \end{itemize}
        \item Specific Humidity Model
        \begin{itemize}
            \item MW: $\texttt{uf}=1 , \texttt{ws}=4320 $
            \item IW: $\texttt{uf}=168$
            \item OPS: $\gamma=1\text{e-5} , \texttt{D}=30 $
        \end{itemize}
        \item Airtight Model
        \begin{itemize}
            \item MW: $\texttt{uf}=2160 , \texttt{ws}=2160 $
            \item IW: $\texttt{uf}=720$
            \item OPS: $\gamma=5\text{e-5} , \texttt{D}=10 $
        \end{itemize}
        \item Wind Speed Model
        \begin{itemize}
            \item MW: $\texttt{uf}=1 , \texttt{ws}=168 $
            \item IW: $\texttt{uf}=24$
            \item OPS: $\gamma=1\text{e-4} , \texttt{D}=10 $
        \end{itemize}
    \end{itemize}
    \item \textbf{Base Classification Model}
    \begin{itemize}
        \item Pressure Model
        \begin{itemize}
            \item MW: $\texttt{uf}=2160 , \texttt{ws}=8640 $
            \item IW: $\texttt{uf}=720$
            \item OPS: $\gamma=5\text{e-5} , \texttt{D}=30 $
        \end{itemize}
        \item Temperature Model
        \begin{itemize}
            \item MW: $\texttt{uf}=1 , \texttt{ws}=4320 $
            \item IW: $\texttt{uf}=168$
            \item OPS: $\gamma=1\text{e-5} , \texttt{D}=150 $
        \end{itemize}
        \item Saturation Vapor Pressure Model
        \begin{itemize}
            \item MW: $\texttt{uf}=336 , \texttt{ws}=4320 $
            \item IW: $\texttt{uf}=720$
            \item OPS: $\gamma=1\text{e-4} , \texttt{D}=30 $
        \end{itemize}
        \item Vapor Pressure Deficit Model
        \begin{itemize}
            \item MW: $\texttt{uf}=1 , \texttt{ws}=168 $
            \item IW: $\texttt{uf}=1$
            \item OPS: $\gamma=1\text{e-5} , \texttt{D}=70 $
        \end{itemize}
        \item Specific Humidity Model
        \begin{itemize}
            \item MW: $\texttt{uf}=1 , \texttt{ws}=2160 $
            \item IW: $\texttt{uf}=2160$
            \item OPS: $\gamma=1\text{e-5} , \texttt{D}=50 $
        \end{itemize}
        \item Airtight Model
        \begin{itemize}
            \item MW: $\texttt{uf}=24 , \texttt{ws}=4320 $
            \item IW: $\texttt{uf}=336$
            \item OPS: $\gamma=1\text{e-3} , \texttt{D}=10 $
        \end{itemize}
        \item Wind Speed Model
        \begin{itemize}
            \item MW: $\texttt{uf}=24 , \texttt{ws}=2160 $
            \item IW: $\texttt{uf}=1$
            \item OPS: $\gamma=1\text{e-5} , \texttt{D}=10 $.
        \end{itemize}
    \end{itemize}
    
\end{itemize}

\subsection{Hyperparameters for Control in Nuclear Fusion Case Study}
The nuclear fusion dataset records measurements in 25 millisecond intervals.
Therefore, in tuning hyperparameters, we design the search grid to represent lengths of time during which evolution of various plasma states are expected to be observable.

For each calibration method, the hyperparameters were tuned by optimizing parity calibration error (PCE, Section~\ref{sec:experiments}) on a validation dataset consisting of 1000 shot's worth of data, over the following grids:
\begin{itemize}
    \item \texttt{uf} $\in [1, 2, 4, 8, 24]$, separately for MW and IW
    \item \texttt{ws} $\in [2, 8, 16, 24, 48, 60, 80, 100, 200]$, for MW
    \item $\gamma$ $\in$ [1e-5, 5e-5, 1e-4, 5e-4, 1e-3, 5e-3, 1e-2], for OPS
    \item \texttt{D} $\in [1, 10, 30, 50, 70, 100, 150, 200]$, for OPS
\end{itemize}

The tuned hyperparameters we used are as follows:
\begin{itemize}
    \item MW: $\texttt{uf}=1, \texttt{ws}=60$
    \item IW: $\texttt{uf}=8$
    \item OPS: $\gamma=5\text{e-3}, \texttt{D}=150$.
\end{itemize}

\newpage
\section{Online Platt Scaling Algorithm}\label{app:ops-algorithm}
\begin{algorithm}[h]
\begin{algorithmic}
	\STATE {\bfseries Input: } $\mathcal{K} = \{(x, y): \norm{(x, y)}_2 \leq 100\}$, time horizon $H$, and initialization parameter $(a_1^\ops, b_1^\ops) = (1, 0) =: \theta_1 \in \mathcal{K}$\;
        \STATE {\bfseries Hyperparameters and default values:} $\gamma = 0.1$, $D = 1$, $A_0 = (1/\gamma D)^2\  \mathbf{I}_2$
    \FOR{$t=1$ {\bfseries to} $H$}
    \STATE Play $\theta_t$, observe log-loss $l(m^{\theta_t}(f(\x_t)), y_t)$ and its gradient $\nabla_t := \nabla_{\theta_t}l(m^{\theta_t}(f(\x_t)), y_t)$
    \STATE $A_t = A_{t-1} + \nabla_t \nabla_t^\intercal$
    \STATE Newton step: $\widetilde{\theta}_{t+1} = \theta_t - \frac{1}{\gamma} A_t^{-1} \nabla_t$
    \STATE Projection: $(a_{t+1}^\ops, b_{t+1}^\ops) = \theta_{t+1} = \argmin_{\theta \in \mathcal{K}} (\widetilde{\theta}_{t+1}-\theta)^{\intercal}A_t(\widetilde{\theta}_{t+1}-\theta)$
    \ENDFOR
    \end{algorithmic}
 	\caption{Online Platt Scaling (based on \citet{gupta2023online})} 
  \label{alg:ops-ons}
\end{algorithm}

\end{document}